\documentclass[12pt]{article}
\usepackage{graphicx}
\usepackage{epsfig}
\usepackage{amsmath,amsthm,amssymb,bm,subfigure,color,hyperref}
\newcommand{\argmax}{\mathop{\rm argmax}\limits}

\begin{document}

\baselineskip 0.7cm

\begin{titlepage}
\renewcommand{\thefootnote}{\fnsymbol{footnote}}
\begin{flushright}
\end{flushright}

\begin{center}
{\large \bf
Outer-Product Hidden Markov Model\\
and Polyphonic MIDI Score Following
}
\vskip 1.2cm
Eita Nakamura$^1$\footnote[1]{Presently with National Institute of Informatics, Tokyo 101-8430, Japan.},
Tomohiko Nakamura$^1$,
Yasuyuki Saito$^2$,\\
Nobutaka Ono$^3$
and Shigeki Sagayama$^1$\footnotemark[1]
\vskip 0.4cm

{\it $^1$  The University of Tokyo,\\
Tokyo 113-0033, Japan\\
$^2$ Kisarazu National College of Technology,\\
Chiba 292-0041, Japan\\
$^3$ National Institute of Informatics,\\
Tokyo 101-8430, Japan}

\vskip 1.5cm
\renewcommand{\thefootnote}{\arabic{footnote}}
\abstract{
We present a polyphonic MIDI score-following algorithm capable of following performances with arbitrary repeats and skips, based on a probabilistic model of musical performances.
It is attractive in practical applications of score following to handle repeats and skips which may be made arbitrarily during performances, but the algorithms previously described in the literature cannot be applied to scores of practical length due to problems with large computational complexity.
We propose a new type of hidden Markov model (HMM) as a performance model which can describe arbitrary repeats and skips including performer tendencies on distributed score positions before and after them, and derive an efficient score-following algorithm that reduces computational complexity without pruning.
A theoretical discussion on how much such information on performer tendencies improves the score-following results is given.
The proposed score-following algorithm also admits performance mistakes and is demonstrated to be effective in practical situations by carrying out evaluations with human performances.
The proposed HMM is potentially valuable for other topics in information processing and we also provide a detailed description of inference algorithms.
}

\vskip 1cm


\end{center}
\end{titlepage}

\setcounter{page}{2}
\baselineskip 0.6cm

\section{Introduction}

Automated matching of notes in musical performances to notes in corresponding scores in real time is called score following, and it is a basic tool for real-time applications such as automatic accompaniment and automatic turning of score pages.
Studies on automatic accompaniment have aimed at realising automatic performances of accompaniments synchronised to human performances based on referential scores.
It is applicable to efficient practices of ensemble music by one or a limited number of performers and opera rehearsals as well as live electronic performances.
Since the first studies on score following and automatic accompaniment \cite{Dannenberg1984,Vercoe1984}, many studies have been carried out on their realisations and improvements (see Ref.~\cite{Orio2003} for a review of studies in this field, and for more recent studies, see {\it e.g.}, Refs.~\cite{Pardo2005,Cont2010,Montecchio2011,Arzt2012}, just to mention a few).

Performers using automatic accompaniments, especially in musical practice, hope to start playing from a score position at will and generally make arbitrary repeats and/or skips (repeats/skips) during performances in practical situations.
Besides such diversity, performers often make mistakes during performances such as errors with pitch and deletions and insertions of notes.
It is desirable for score-following algorithms to handle these repeats/skips as well as performance mistakes.
Score-following algorithms handling repeats/skips have been studied in Refs.~\cite{Tekin2004,Pardo2005,Oshima2005,Tekin2006,Takeda2006MUS}, and more recently in Ref.~\cite{Montecchio2011}.
In the earlier studies, repeats/skips to limited score positions have been discussed, but their capabilities of following repeats/skips to unspecified score positions have not been guaranteed.
Also, the algorithms for musical instrument digital interface (MIDI) performances have essentially been limited to monophonic cases \cite{Tekin2004,Pardo2005,Oshima2005}.

When these algorithms are extended to handling arbitrary repeats/skips, their computational complexity increases significantly for longer scores \cite{Takeda2006MUS,Tekin2006}.
Most score-following algorithms including those mentioned above use either dynamic time warping (DTW) \cite{Dannenberg1984,Bloch1985} or hidden Markov models (HMMs) \cite{Cano1999} and apply dynamic programming (DP) to reducing the computational complexity required for real-time processing.
As we will discuss later in detail, computational complexity increases in square order of the number of notes in scores when we simply extend their algorithms to handling arbitrary repeats/skips, and it is hard to process longer scores that commonly appear in classical music.
The case is similar if we restrict repeats/skips to limited score positions since the number of these score positions usually increases with the length of the score.

In order to solve the problem, the author in Ref.~\cite{Tekin2006} has proposed a score-following algorithm based on a multi-agent system which can handle polyphonic performances with arbitrary repeats/skips as well as performance mistakes.
In the thesis, the author introduces multiple agents each corresponding to a score position, and uses heuristics for calculating their activations, which are then used as inputs for the locator agent that detects the current score position.
The author discusses that the algorithm could follow simulated performances with large repeats/skips with an acceptable level of success.
Since the optimality of the algorithm is not ensured\footnote{In the discussion involving Fig.~3.9 in Ref.~\cite{Tekin2006}, it is argued that ``non-regional activities'' can occur in addition to ``regional activities'' which are indications of approximate score positions.
Although it seems that the method to extract the explained regional activities and its accuracies are crucial for the effectiveness of the score-following algorithm, no explanations on these issues are given.}, its effectiveness should at least be confirmed empirically by systematic evaluations.
However, neither systematic evaluations using real human performances nor comparisons with the earlier algorithms were done in the work.
Moreover, details of the algorithm are not fully expanded, which makes it virtually impossible to reproduce the results reported or to examine the algorithm further.

Meanwhile, authors in Ref.~\cite{Takeda2006MUS} have extended the earlier algorithms using the HMM to handling polyphonic performances with arbitrary repeats/skips, which is affected by the problem of large computational complexity.
Recently, authors in Ref.~\cite{Nakamura2013} proposed a score-following algorithm using a constrained type of HMM, which significantly reduced the computational complexity.
It was found that if the probability of making repeats/skips in the HMM are uniformly distributed over all score positions, the computational complexity for score following can be reduced to linear order without pruning, so that arbitrary repeats/skips can be handled for much longer scores.
The score-following algorithm was capable of following arbitrary repeats/skips within about three chords (depending on scores) after performances were resumed, and much faster than the earlier algorithms without modeling arbitrary repeats/skips.

The score positions where performers resume after repeats/skips are not completely random in actual performances and are affected by the purpose of repeats/skips and performers' understanding of musical structures, {\it i.e.} forms and phrase structures as well as beat structures such as bars.
There may also be tendencies on score positions where they stop playing before making repeats/skips.
Once such knowledge on the tendencies of repeats/skips is given in advance, the accuracy of score following and required time for following repeats/skips after performances are resumed ---called the following time in the rest of this paper--- are expected to be improved.
Whether this is possible without causing significant increases in computational complexity and how much the results are improved is not clear.
Since the quality of automatic accompaniment and other applications of score following is crucially dependent on the accuracy and the following time, it is important to improve them and to quantify their limitations in principle.

In this paper, we demonstrate that such knowledge can be incorporated in the performance model in a stochastic manner by using a new type of HMM and derive an efficient score-following algorithm with linear-order complexity.
Since HMM is a basic tool that is widely used for information processing and the phenomenon of large repeats/skips is also seen in speech and other human actions, the proposed HMM is potentially valuable for other applications, and we discuss efficient inference algorithms for it in detail.
The proposed HMM is a generalisation of that proposed in Ref.~\cite{Nakamura2013}.
Since this related work has not been published in English, the main results are reviewed in this paper.
We provide a theoretical discussion on the degree of improvement to the score-following results.
We also evaluate the score-following algorithm using human performances.

Score-following algorithms generally accept either acoustic signals or symbolic MIDI signals of performances as input.
Score-following algorithms for acoustic signals have been improved over the years and fairly high accuracies have been reported even for polyphonic performances including piano and orchestral music, at least for clean performances as in recordings (see {\it e.g.}, Refs.~\cite{Raphael1999,Raphael2011,Cont2010,Arzt2012}).
Using acoustic inputs certainly has advantages in being applicable to a wider range of instruments and situations.
Drawbacks are latencies of order 50--100 ms in detecting onsets, which can be significant in ensemble music but are hard to avoid in principle, and that the quality can be influenced by background noise and other acoustics, especially for automatic accompaniment, and also by calibrations of microphones\footnote{There are also reports for possible problems in sudden tempo changes, very fast notes, extensive use of pedals in pianos and vibraphones \cite{Schwarz2004,Lemouton2009}.
Since these reports do not perform quantitative tests, systematic evaluations for these effects are in order.}.
On the other hand, using MIDI inputs has advantages in quick correspondences to onsets and in clean signals, which do not suffer from noise, resonances, and any background acoustics including pedal effects.
This motivates uses of MIDI inputs in certain situations because of the potentially vast demand for score following of polyphonic keyboard performances.
For the reason that MIDI signals (discrete in pitches and continuous in time) and acoustic signals (frame-wise discrete in time and continuous in features) are different in nature, and separate discussions are required, we focus on polyphonic MIDI signals for input performance signals in this paper.
A similar score-following algorithm for acoustic input allowing arbitrary repeats/skips is discussed elsewhere \cite{tNakamura2013}.

\section{Probabilistic performance model}\label{sec:PerfmHMM}
\subsection{Performance uncertainty and its probabilistic modeling}\label{sec:PerformanceUncertainty}
In order to keep synchronised with a human performance, it is necessary to align score positions to the performance and to estimate the tempo.
Score following is generically a challenging problem to solve because human performances vary widely even if they are based on the same score.
This fact is often called uncertainty or indeterminacy in musical performances.
Six typical examples important for our purposes are listed below.

\begin{itemize}
\item[a)] {\bf Tempos.} In many musical scores, performers determine absolute values of tempos and the ways they are varied.
A fermata may also be regarded as a local variation of a tempo, whose realisation is up to performers.

\item[b)] {\bf Small fluctuations in onset times.} According to musical intention, technical constraints of performers and/or physical constraints of instruments, performed onset times have small fluctuations, some of which may be interpreted as tempo variations. 

\item[c)] {\bf Dynamics and articulations.} Although dynamics and articulations are usually notated explicitly, they are often added or modified by improvisation.
In either case, the details of their realisations are determined by performers.

\item[d)] {\bf Performance mistakes.} Mistakes in performances can also be interpreted as performance uncertainties.
They derive from constraints on performers' skills or misreadings of the score and they result in pitch errors, note deletions and insertions, and duration/articulation errors, etc.

\item[e)] {\bf Ornaments.} Realisations of ornaments depend on performers.
For example, the number of notes in a trill and its rapidity are determined by performers' skill and convention and also by chance. 
Realisations of arpeggios and grace notes also depend on performers.

\item[f)] {\bf Repeats and skips.} Repetitions of phrases or skips to remote score positions may be made, especially in practices.
Deletions or additions of a repeated section are also common in concert performances, not to mention performances in open forms.
\end{itemize}

These six uncertainties provide both difficulty and motivation to score following.
It is essential for automatic accompaniment to accommodate these uncertainty while extracting and/or estimating necessary information to be reflected in the accompaniment.

Our approach to score following is that the above uncertainties can be interpreted as being stochastically generated, and the act of performance is described with a probabilistic generative model.
With this generative modeling, score following can be treated as solving the inverse problem of the generating process.
The idea of using probabilistic models for score following is not new and has in fact become a major idea since its first appearance in Ref.~\cite{Cano1999} (see also Ref.~\cite{Orio2003}).
Our main contribution in this respect is to extend the performance model to incorporate the repeats and skips (f) in the above list.
We also discuss polyphonic MIDI performance models \cite{Bloch1985,Schwarz2004} which generalise the monophonic case discussed in Ref.~\cite{Pardo2005}.
In the rest of the paper, we focus on the score-position estimation problem.
For the estimation of tempo, we refer the reader to Refs.~\cite{Raphael2001,Cemgil2001,Cemgil2003,Cont2010}.

\subsection{Properties of repeats/skips in performances during practice}\label{sec:RepeatSkip}
There are several causes or purposes of repeats/skips in performances, especially during practices.
A common cause is due to correcting for mistakes in performances, which usually results in skipping backward to a comparably close score position, through a few notes/chords to a few bars.
A similar cause is refreshment of performances which may occur at any score positions.
Another cause is due to practicing a specific phrase or section efficiently in order to remember the score and performance details and/or to polish up the performance. 
In this case, performers often repeat the phrase or section many times.
Forward skips are made when performers want to omit phases or sections to efficiently practice longer pieces or because of some musical demand.
From now on, we will use the word ``skip'' to mean both a repeat, or a backward skip, and a forward skip for the sake of simplicity if there is no ambiguity.

The score positions where performers stop before skips and where they resume performing are supposed to depend on the cause and purpose of the skips and also on their understanding of the musical structure.
We call these score positions ``stop positions'' (where they stop before making skips) and ``resumption positions'' (where they resume performing) in what follows.
Performers are likely to stop at any score positions when simply correcting mistakes and resume from a few notes/chords backward or from the beginning of the bar or phrase.
When repeating or skipping sections, performers are more likely to resume from the beginning of a section.

\begin{figure}[tbp]
\begin{center}
\includegraphics[clip,width=0.7\columnwidth]{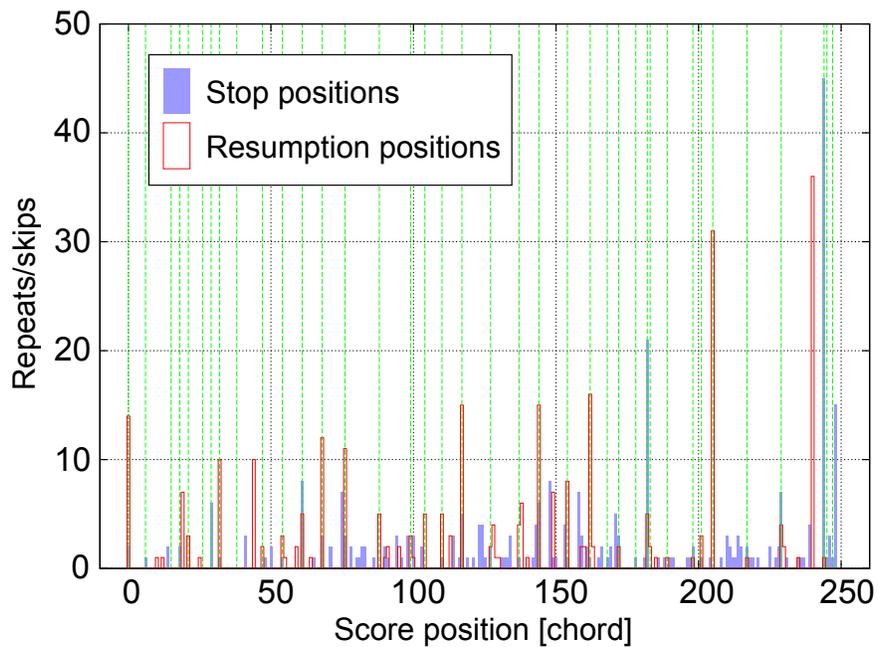}
\end{center}
\caption{Distribution of stop and resumption positions in the performances during practicing Debussy's ``La fille aux cheveux de lin.'' The data were extracted from performances by three pianists (see Sec.\ \ref{sec:Evaluation} for details). The score positions indicate how many chords precede corresponding notes and vertical dashed lines indicate downbeats of bars.}
\label{fig:SRPositionsDeb}
\end{figure}
\begin{figure}[tbp]
\begin{center}
\includegraphics[clip,width=0.7\columnwidth]{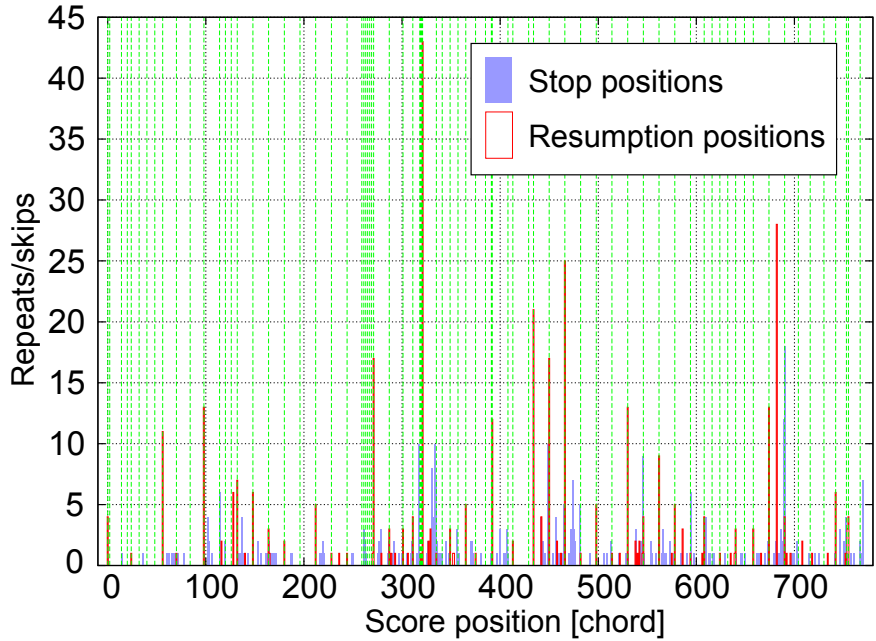}
\end{center}
\caption{Distribution of stop and resumption positions in the performances during practicing the exposition of first movement from Mozart's sonata for two pianos (the first piano part). See caption in Fig.\ \ref{fig:SRPositionsDeb}.}
\label{fig:SRPositionsMoz}
\end{figure}
\begin{figure}[tbp]
\begin{center}
\includegraphics[clip,width=0.7\columnwidth]{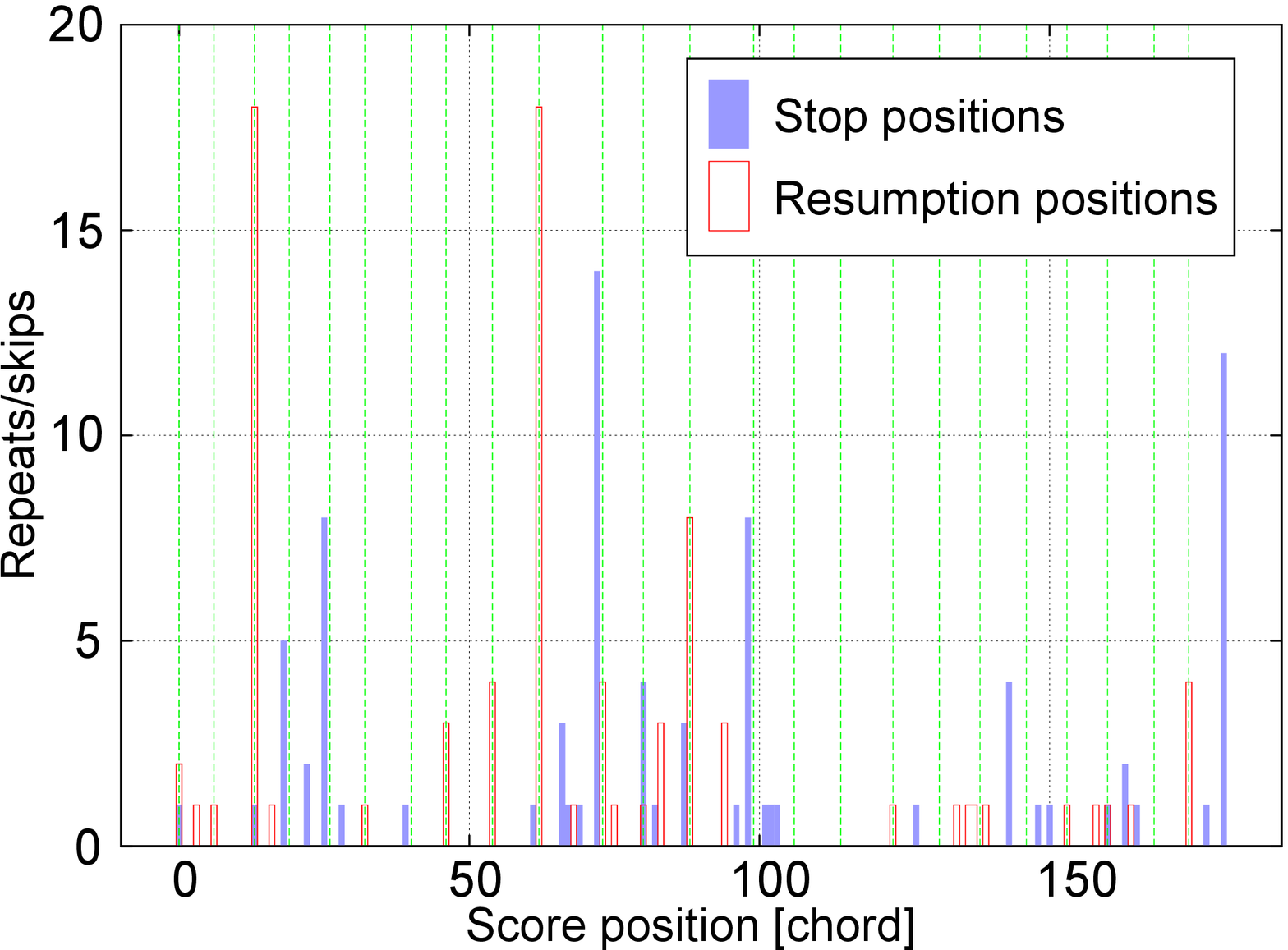}
\end{center}
\caption{Distribution of stop and resumption positions in the performances during practicing Mussorgsky's ``Promenade.'' See caption in Fig.\ \ref{fig:SRPositionsDeb}.}
\label{fig:SRPositionsMus}
\end{figure}
In Figs.\ \ref{fig:SRPositionsDeb}, \ref{fig:SRPositionsMoz} and \ref{fig:SRPositionsMus}, the distributions of stop and resumption positions taken from human performances recorded during practice are shown for three different piano pieces, {\it i.e.}, Debussy's ``La fille aux cheveux de lin,'' the exposition of the first movement from Mozart's sonata for two pianos in D major (the first piano part) and Mussorgsky's ``Promenade'' (the first piece from the suite ``Pictures at an Exhibition'' for piano).
Three pianists played the same pieces, and the stop and resumption positions were manually analysed (for more details, see Sec.\ \ref{sec:Evaluation}).
The horizontal axis indicates the score positions in terms of chords\footnote{Henceforth, a ``chord'' within our context means a set of one or multiple notes performed (almost) simultaneously.} and the filled and lined histogram shows the number of repeats/skips with the stop and resumption position at the corresponding score positions out of 288 (resp.~373, 83) repeats/skips in total for the piece by Debussy (resp.~Mozart, Mussorgsky).
The dashed vertical lines indicate the positions of bar onsets, or alternatively the positions of the last chord in the previous bar if there is no chord on the bar onset, and of the first chord in the bar further if there are no chords in the previous bar.

Although the histograms are somewhat sparse, they clearly reveal the performers' tendencies in stop and resumption positions.
By looking at them more closely, we see the resumption positions are more likely to occur at bar onsets and beginnings of phrases.
For example, the most frequent resumption position for Debussy's piece is at the 240th chord, or the arpeggio in bar 36 \footnote{The bar numbers both here and in what follows are given for the convenience of readers. See, for example, the first edition of Debussy's Pr\'eludes Premier Livre published by Durand \& Cie.}.
The high frequencies of resumption and stop positions at the 244th chord are actually accidental, since this is the result of repeated realisations of the arpeggio, which is represented as a sequence of five chords according to our convention (see footnote \ref{footnote:Ornaments}), and these are more appropriately regarded as insertions (see the following sections).
The next five most frequent resumption positions at the 0th, 117th, 144th, 162nd and 205th chord are all on bar onsets at bars 1, 19, 22, 24 and 33, and those except for the 144th chord are the beginning of a phrase with an indicated tempo change.
Also note that there are less frequent skips with resumption positions in the middle of bars.
They occur if a phrase begins with an upbeat or when performers repeat or skip chords to suddenly correct for mistakes or simply to refresh the performance.
Some of these resumption positions are hard to predict as they reflect performers' understanding of musical structure and feeling, and they are distributed quite arbitrarily in effect.
The stop positions also have a tendency to distribute more frequently at boundaries of bars and phrases, but the distribution is more widespread than that of resumption positions.
In Fig.\ \ref{fig:SRPositionsDeb}, high frequencies can be seen at the 182nd chord, or at the beginning of bar 28, where a phrase ends, and at the end of the piece.
However, other stop positions are in the middle of bars and are not necessarily associated with musical structures.
Similar analyses can be done for the other two pieces.

The results indicate that performers stop and resume frequently at specific score positions which are typically at boundaries of bars, phrases and sections, and they less frequently stop and resume at other score positions which are hard to expect.
Therefore, it is necessary in score following to handle skips with arbitrary stop and resumption positions to avoid getting lost by skips involving unexpected score positions.
Once the tendencies of distributed stop and resumption positions are given in advance, we can expect to obtain better results for score following by using this information since score positions can be more efficiently searched.
We will return to this point in Sec.\ \ref{sec:DiscOnImprovement} after discussing explicit models and algorithms.

\subsection{Performance HMM}
In the rest of this section, we discuss our performance model based on an HMM.
The process of performance can be described with a state space model, whose state corresponds to a musical event, or a chord as defined in Sec.~\ref{sec:RepeatSkip}, in the performance score\footnote{\label{footnote:Ornaments}
Although other types of musical events like ornaments ({\it e.g.}\ trills and arpeggios) are important for score following \cite{Dannenberg1988,Tekin2006,Cont2010}, we confine ourselves to ordinary chords and treat ornaments as explicit realisations in this paper (see later discussion).}.
The performance score $c_{1:N}=(c_i)_{i=1}^N$ is described as a sequence of $N$ chords denoted by $c_i$'s.
As the performance continues, the performer scans the score from chord to chord, and this process is modeled as a sequence of stochastic variables $I_{1:M}=(I_m)_{m=1}^M$, where $M$ is the number of performed notes and each $I_m$ ($m=1,\cdots,M$) takes values in $1,\cdots,N$.
The probability of $I_m$ is supposed to depend on the previous performed events $I_1,\cdots,I_{m-1}$.
If we further assume that the process is Markovian, the probability of $I_{1:m}$ is written as
\begin{equation}\label{eq:prob_PerfmEvt}
P(I_{1:m})=P(I_1,\cdots,I_m)=\prod_{m'=1}^mP(I_{m'}|I_{m'-1})
\end{equation}
with the understanding of notation for the initial probability $P(I_1|I_0)\equiv P(I_1)$.

In actual performances, the performed chords introduced above themselves are not directly observed, and what are observed are resulting performed notes.
For example, a note in $c_i$ with correct pitch G\#4 may be observed as a G4 if there is pitch error in the performance, but still the intention of the performer is to perform chord $c_i$.
To distinguish between these intended chords from the actual observations, we call the latter observed events, which constitute the performance signal.
We obtain information on pitch, velocity, onset time and the released time of each performed note from the performance signal in MIDI format.
Of these, the released time is rather uncertain due to individual performer skills and articulations and, in piano performance with pedals,  it does not always match the damping time \cite{Schwarz2004}.
Velocity is also largely uncertain and not supposed to be important for estimating score-position in typical scores.
In the rest of this paper, we only use onset events for score following and the pitch of an observed event is denoted by $o_m$ and its onset time by $t_m$.

Because an observed event resulting from a performed event has uncertainty caused, for example, by performers' skills and physical constraints, it can also be described stochastically.
Assuming that an observed event $O_m$ of an intended chord $I_m$ is only dependent on $I_m$, the probability of $O_m$ is given as a conditional probability
\begin{equation}\label{eq:prob_ObsEvt}
P(O_m|I_m).
\end{equation}
Eq.~(\ref{eq:prob_ObsEvt}) together with Eq.~(\ref{eq:prob_PerfmEvt}) shows that our performance model has the form of an HMM \cite{Rabiner1989}. We call it the performance HMM.
Following traditional notations, we denote the transition and output probabilities by
\begin{align}
P(I_m=j|I_{m-1}=i)&=a_{i,j},\label{eq:transP}\\
P(O_m=o|I_{m}=i)&=b_i(o) \label{eq:emitP}
\end{align}
in the following. The joint probability of performed events and observed events can be written with the notation as
\begin{equation}\label{eq:prob_PerfmObs}
P(I_{1:m}=i_{1:m},O_{1:m}=o_{1:m})=\prod_{m'=1}^m a_{i_{m'-1},i_{m'}}b_{i_{m'}}(o_{m'}),
\end{equation}
where by abuse of notation, we again understand $a_{i_0,i_1}=P(I_1=i_1)$ as the initial probability.

\subsection{Description of chords in the performance HMM}
\begin{figure}[tb]
\begin{center}
\includegraphics[clip,width=0.7\columnwidth]{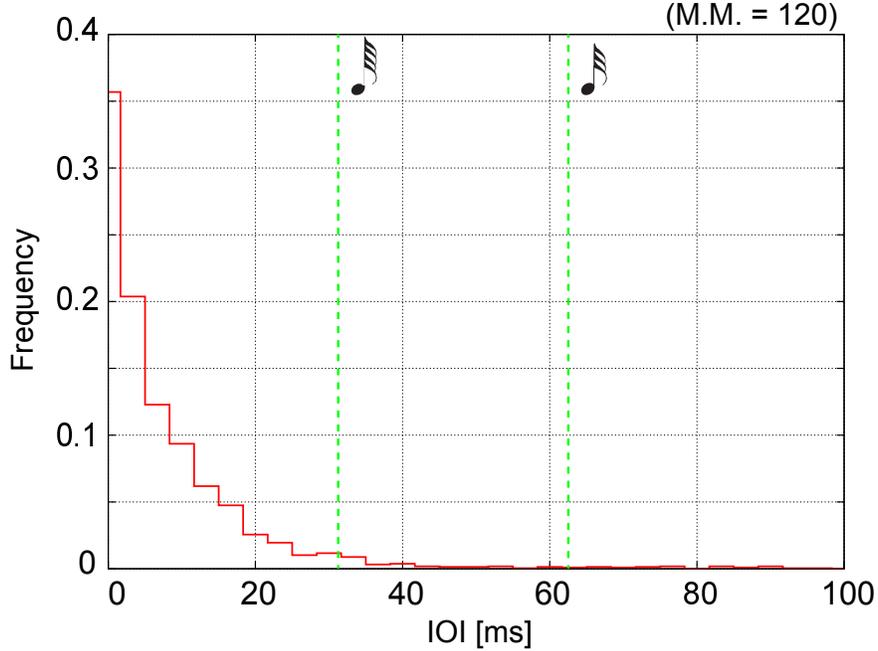}
\end{center}
\caption{Distribution of IOIs between chordal notes. The data were extracted from performances described in Sec.\ \ref{sec:Evaluation}. Sojourn times of 32nd and 64th notes in tempo M.M. $=120$ are indicated by vertical dashed lines for illustration.}
\label{fig:ChordIoi}
\end{figure}
Multiple notes in a chord are written simultaneously in scores but actually performed at different times in indeterminate order.
These notes are described as outputs of an HMM state through self transitions.
Fig.\ \ref{fig:ChordIoi} shows the distribution of inter-onset intervals (IOIs) between notes in a chord taken from human performances (we have used the same data as those described in Sec.\ \ref{sec:Evaluation}), and we can see that the IOIs spread out within about 40 ms.
A similar value is obtained by measurements of expressive performances \cite{Repp1996,Goebl2001}.
Since this is much smaller than IOIs between chords in typical scores\footnote{ For example, the duration of a 64th note in tempo M.M. $=120$ is about 30 ms, see Fig.\ \ref{fig:ChordIoi}.}, a threshold of $\Delta t_{\rm limit}=35$ ms is set and we approximate that the state transitions between notes with IOIs less than $\Delta t_{\rm limit}$ only occur as self transitions:
\begin{equation}
a_{i_{m-1},i_m}=\delta_{i_{m-1},i_m}\quad\text{for $t_m-t_{m-1}<\Delta t_{\rm limit}$},
\end{equation}
where $\delta_{i,j}$ denotes Kronecker's delta.
Note that the existence of a threshold does not restrict larger IOIs between chordal notes because they are treated as chord insertions, which is described in the next section.

If a pitch $o$ is contained in a chord $c_{i}$, the output probability $b_{i}(o)$ is positive.
Because pitch errors are represented by emissions of notes not contained in the intended chord, we have small but positive values of $b_{i}(o)$ for other pitches $o$.

\subsection{Transition probability}
The transition probability $a_{i,j}$ probabilistically represents the order of performed chords.
For a monophonic performance without note insertions or deletions, $a_{i_{m-1},i_m}=\delta_{i_{m-1}+1,i_m}$.
A chord/note insertion and deletion of a chord are described as a self transition and a small skip transition, which correspond to $i_m=i_{m-1}$ and $i_m=i_{m-1}+2$ in $a_{i_{m-1},i_m}$.
All these state transitions are between neighbouring states and have been treated in preceding studies including Refs.~\cite{Bloch1985,Schwarz2004}.
As we discussed in the previous section, the performance of a chord is represented by a succession of self transitions with small IOIs.

\begin{figure}[tb]
\begin{center}
\includegraphics[clip,width=1\columnwidth]{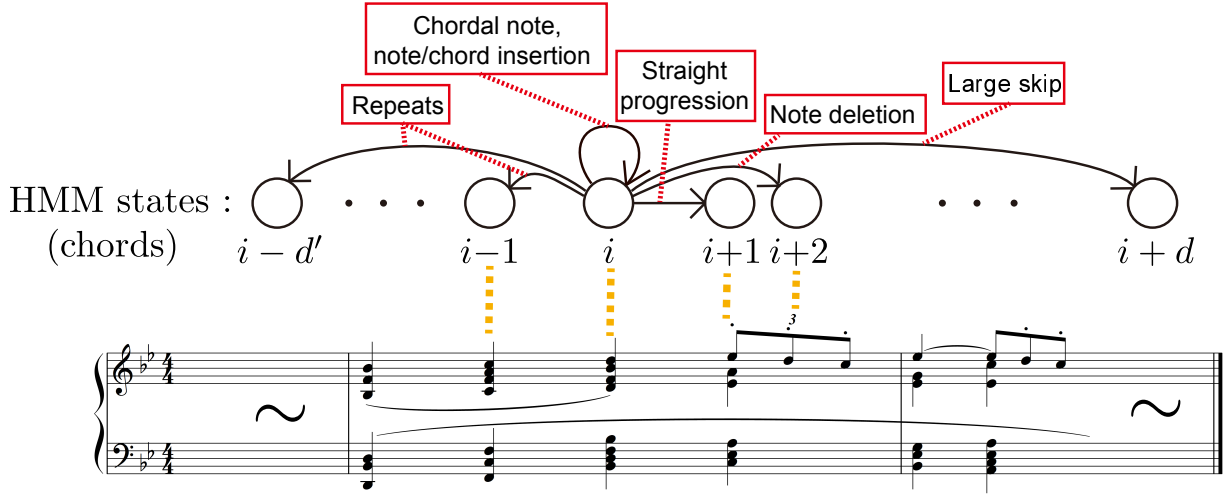}
\caption{Topology of state transition probability for the performance HMM. $i$'th state corresponds to $i$'th chord of performance score.}
\label{fig:HMMTopology}
\end{center}
\end{figure}

To describe large skips in performances, transition probabilities $a_{i,j}$ with large $|j-i|$ must be considered \cite{Pardo2005}.
In order to model arbitrary skips, the performance HMM must have transitions connecting two arbitrary states which extends the model in Ref.~\cite{Pardo2005}.
The state-transition topology of our performance HMM is illustrated in Fig.~\ref{fig:HMMTopology}.
%

\section{Outer-product HMM and score following}\label{sec:Model}
\subsection{Estimation of score position and computational complexity}
Given a generative model of musical performances, {\it i.e.}, the performance HMM, the estimation of score positions can be treated as a probabilistic inverse problem.
This is done by calculating the most likely state given observation events.
Because score following involves real-time online matching, we can only use past observation events.
An online version of the Viterbi algorithm (estimating the most likely state sequence up to the present) or the forward algorithm (estimating the most likely present state) can be exploited for this purpose.
In Ref.~\cite{Pardo2005}, both algorithms were compared and they found similar results with the Viterbi algorithm slightly outperforming the forward algorithm.
Although we mainly discuss the Viterbi algorithm in the following, a similar discussion is also valid for the forward algorithm.

Provided observation events $o_{1:M}$, we can obtain the most likely state sequence $\hat{i}_{1:M}$ by calculating
\begin{equation}
\argmax_{i_1,\cdots,i_M}P({i_{1:M}}|o_{1:M})
=\argmax_{i_1,\cdots,i_M}P(o_{1:M}|{i_{1:M}})P({i_{1:M}})
=\argmax_{i_1,\cdots,i_M}\left[\prod_{m=1}^Ma_{i_{m-1},i_m}b_{i_m}(o_m)\right],
\end{equation}
where the first equation follows from Bayes' theorem.
This can be iteratively obtained with the online version of the Viterbi algorithm.
If the maximum likelihood for state $i_{M-1}$ at time $t_{M-1}$ is given as
\begin{equation}
\hat{p}_{M-1}(i_{M-1})=\max_{i_1,\cdots,i_{M-2}}\left[\prod_{m=1}^{M-1}a_{i_{m-1},i_m}b_{i_m}(o_m)\right],
\end{equation}
one can update the value at time $t_M$ given observation event $o_M$ as
\begin{equation}\label{eq:ViterbiUpdate}
\hat{p}_M(i_M)=\max_{i_{M-1}}\left[\hat{p}_{M-1}(i_{M-1})a_{i_{M-1},i_M}b_{i_M}(o_M)\right].
\end{equation}
The most likely state $\hat{i}_M$ (the latest one in the most likely state sequence) at time $t_M$ is then given as $\hat{i}_M=\argmax_{i_M}\hat{p}_M({i_M})$.
For score following to work in real time, it is necessary to update Eq.\ (\ref{eq:ViterbiUpdate}) within a sufficiently short time.
Without incorporating large skips, $a_{i,j}=0$ for $i$ and $j$ satisfying $j<i-D_1$ or $j>i+D_2$ for small $D_1$ and $D_2$, and evaluation of the right-hand side of Eq.\ (\ref{eq:ViterbiUpdate}) requires ${\cal O}(D)$ computations of probability, where $D=D_1+D_2+1$, and thus the total computational complexity of updating Eq.\ (\ref{eq:ViterbiUpdate}) is ${\cal O}(DN)$, where $N$ is the number of chords in the referential performance score.
In contrast, $a_{i,j}\neq0$ for all $i$ and $j$ if we allow arbitrary skips, and then the computational complexity of updating Eq.\ (\ref{eq:ViterbiUpdate}) is ${\cal O}(N^2)$.
As we will quantitatively demonstrate in Sec.\ \ref{sec:EvalCompComplex}, this ${\cal O}(N^2)$ complexity is too large to process scores of a practical length\footnote{Most classical musical pieces have ${\cal O}(100$--$10000)$ chords. For example, the solo piano part of Rachmaninoff's piano concerto No.\ 3 d-moll has $N\simeq5000$ chords only in the first movement.} without serious time delays.
The ${\cal O}(N^2)$ complexity is typical for other DP techniques such as the forward algorithm and the DTW method, and it is crucial to reduce the computational complexity for practical applications of score following for performances with skips.
We also need to remark that a similar situation occurs if we limit the resumption positions to specific locations such as the first beats of each bar, because the number of these resumption positions usually increases with (and probably in proportional to) the length of the score.
Otherwise we can only have fewer resumption positions and this would lead to poorer accuracy and a late following time of score following.

\subsection{Performance HMM with uniform repeat/skip probability and fast Viterbi algorithm}
\label{sec:UniformModel}

Although one might consider some pruning techniques to reduce computational complexity, pruning is not valid within the context of handling arbitrary skips since skips rarely occur compared to other state transitions.
Therefore, it seems necessary to introduce some constraints to the performance HMM.

The problem with large computational complexity arises from the nonzero values of the transition probability $a_{i,j}$ for large $|i-j|$.
In Ref.~\cite{Nakamura2013}, we demonstrated that under an assumption of uniform skip probability in the performance HMM, the online Viterbi algorithm can be refined and computational complexity can be reduced.
Although this assumption is crude, it is pragmatic in situations where prior knowledge of musical structures such as bars or sections is not given, {\it e.g.}, as in MIDI files, or when a variety of performers are expected.

Here, we briefly describe the main results in Ref.~\cite{Nakamura2013}.
The above assumption on the transition probability is summarised as
\begin{equation}
a_{i,j}=\gamma,\quad \text{for $j<i-D_1$ or $j>i+D_2$}
\end{equation}
where $D_1$ and $D_2$ are again small positive integers which define a neighbourhood of states.
Since skips rarely occur compared to transitions to neighbouring states, we can assume $\gamma$ is smaller than $a_{i,j}$
with $i-D_1\leq j\leq i+D_2$.
In this type of HMM, the right-hand side of Eq.\ (\ref{eq:ViterbiUpdate}) is given as
\begin{align}
\hat{p}_M(i_M)&=\max_{i_{M-1}}\left[\hat{p}_{M-1}(i_{M-1})a_{i_{M-1},i_M}b_{i_M}(o_M)\right]\\
&=b_{i_M}(o_M)\max\left\{\max_{i_{M-1}\in{\rm nbh}(i_M)}\left[\hat{p}_{M-1}(i_{M-1})a_{i_{M-1},i_M}\right],
\gamma\max_{i_{M-1}}\left[\hat{p}_{M-1}(i_{M-1})\right]\right\},
\label{eq:FastViterbiUniform}
\end{align}
where ${\rm nbh}(i_M)=\{j|j-D_1\leq i_M\leq j+D_2\}$ denotes the set of neighbouring states of $i_M$.
Note that $\displaystyle\max_{i_{M-1}}\,[\hat{p}_{M-1}(i_{M-1})]$ is calculated in the previous Viterbi update.
The computational complexity is now reduced to ${\cal O}(DN)$ and it is much smaller than ${\cal O}(N^2)$ if $D\ll N$.
We refer to the above HMM as the uniform skip model in the following.

\subsection{Outer-product HMM}
As we discussed in Sec.\ \ref{sec:PerformanceUncertainty}, the stop and resumption positions of skips have a certain tendency in actual performances.
Let us now discuss how we can incorporate this knowledge into the performance model.
In general, such knowledge can be incorporated in the transition probability matrix $a_{i,j}$.
As we previously explained above, however, the computational complexity of score following is too large for practical use for a generic transition probability matrix with skips.
In the following, we will argue that the computational complexity can be reduced similarly as above with a simplifying assumption on the form of the transition matrix, or equivalently on the distribution of stop and resumption positions for large skips.

If we assume that the distribution of stop positions and that of resumption positions are independent, the transition probability matrix can be written as
\begin{equation}\label{eq:OuterProductTrMatrix}
a_{i,j}=\alpha_{i,j}+S_ir_j=\alpha_{i,j}+N\bar{\gamma}s_ir_j,
\end{equation}
where $\alpha_{i,j}$ is a band matrix satisfying $\alpha_{i,j}=0$ unless $i-D_1\leq j\leq i+D_2$, which describes transitions within neighbouring states.
Here, $S_i$ corresponds to the probability of making skips at state $i$ and $\bm r=(r_j)$ is the distribution of resumption positions.
From the normalisation condition, we have $1=\sum_ja_{i,j}=\sum_j\alpha_{i,j}+S_i\sum_jr_j$ for any $i$.
Without loss of generality, we can assume $\sum_jr_j=1$ and then we have $S_i=1-\sum_j\alpha_{i,j}$.
If we define $\bar{\gamma}\equiv\sum_iS_i/N$ and $s_i\equiv S_i/(N\bar{\gamma})$, the second equation in Eq.~(\ref{eq:OuterProductTrMatrix}) holds with $\sum_is_i=1$.
The quantity $\bar{\gamma}$ is interpreted as the averaged probability of making skips and when it is sufficiently small, $\bm s=(s_i)$ approximates the distribution of stop positions.
Since the probability of making skips is small, $a_{i,j}\simeq\alpha_{i,j}$ holds for $i-D_1\leq j\leq i+D_2$.
In particular, we can assume $\alpha_{i,j}\geq 0$ in practice.

Although the assumption in Eq.\ (\ref{eq:OuterProductTrMatrix}) was mainly introduced from the requirement to reduce the computational complexity of the inference algorithm, it is not easy to obtain the full transition probability matrix $a_{i,j}$ for all $i$ and $j$ without such an assumption.
This is because we need a method of estimating the transition probability from the score information with sufficient accuracy or a tremendous amount of performance data for training for this purpose, both of which are not available at the moment.
How to obtain the distributions, $\bm s$ and $\bm r$, is also a problem in this reduced model.
We will discuss this problem along with possible deviations in the actual transition probability from the reduced one in Sec.\ \ref{sec:DiscEvalutionResult}.

In the next section, we derive efficient inference algorithms for HMMs with a transition matrix of the form given in Eq.\ (\ref{eq:OuterProductTrMatrix}).
As shown in the appendix, the result remains valid if we generalise the output probability in Eq.\ (\ref{eq:emitP}) to a Mealy-type output probability matrix which satisfies a similar constraint as $a_{i,j}$ as
\begin{equation}\label{eq:OuterProductOutputMatrix}
b_{i,j}(o)=P(O_m=o|I_{m-1}=i, I_{m}=j)=
\begin{cases}
\beta_{i,j}(o), & i-D_1\leq j\leq i+D_2;\\
v_i(o)u_j(o), &{\rm otherwise}.
\end{cases}
\end{equation}
For example, this form of output probabilities is useful in constructing a score-following algorithm that allows any transpositions in performances, for which intervals between successive pitches instead of pitches themselves are used as observations and the output probability depends on both states before and after transitions.
In this case, outputs for transitions inside a chord or to the next chord should be appropriately described by $\beta_{i,j}(o)$ and outputs for large skips can be set, for example, as $v_i(o)=1$ and $u_j(o)=$ ``the overall probability of intervals $o$ for skips to state $j$''.
Since the essential assumption in Eqs.\ (\ref{eq:OuterProductTrMatrix}) and (\ref{eq:OuterProductOutputMatrix}) is that $a_{i,j}$ and $b_{i,j}(o)$ are represented as an outer product of vectors for $i$ and $j$ with large $|i-j|$, we call the model an outer-product HMM.
Since the uniform skip model described in the previous section is obtained by setting $S_i=N\gamma$, $r_j=1/N$, $\beta_{i,j}=b_j(o)$, $v_i(o)=1$ and $u_j(o)=b_j(o)$ in Eqs.\ (\ref{eq:OuterProductTrMatrix}) and (\ref{eq:OuterProductOutputMatrix}), the outer-product HMM is a generalisation of the uniform skip model with relation $\gamma=\bar{\gamma}/N$.

\subsection{Inference algorithms for the outer-product HMM}
\label{sec:InferenceAlg}
We now derive an efficient algorithm for the Viterbi update in Eq.\ (\ref{eq:ViterbiUpdate}), which can be rewritten for simplicity as
\begin{equation}\label{eq:OuterProductViterbiUpdate}
\hat{p}_M(i)=\max_{j}\left[\hat{p}_{M-1}(j)a_{j,i}b_i(o_M)\right].
\end{equation}
For a general case with output probability given as in Eq.~(\ref{eq:OuterProductOutputMatrix}), see Appendix \ref{app:InferenceAlgorithmForGeneralCase}.
In the following, we assume $\alpha_{j,i}\geq 0$.
While it is a natural assumption for performance HMM, how we can relax this assumption is also discussed in Appendix \ref{app:InferenceAlgorithmForGeneralCase}.
Substituting Eq.~(\ref{eq:OuterProductTrMatrix}) and using $\alpha_{j,i}\geq 0$, we can calculate this as
\begin{align}
\hat{p}_M(i)
&=b_i(o_M)\max\left\{\max_{j \in {\rm nbh}(i)}[\hat{p}_{M-1}(j)a_{j,i}], \max_{j}[\hat{p}_{M-1}(j)S_jr_i]\right\}\\
&=b_i(o_M)\max\left\{\max_{j \in {\rm nbh}(i)}[\hat{p}_{M-1}(j)a_{j,i}], r_i\max_{j}[\hat{p}_{M-1}(j)S_j]\right\},
\label{eq:FastViterbiOuterProduct}
\end{align}
where ${\rm nbh}(i)=\{j|j-D_1\leq i\leq j+D_2\}$ again denotes the set of neighbouring states of $i$.
Since the factor $\max_{j}[\hat{p}_{M-1}(j)S_j]$ in the last equation is independent of $i$ and can be calculated with ${\cal O}(N)$ complexity, the last expression in Eq.\ (\ref{eq:FastViterbiOuterProduct}) for all $i$ can be evaluated with ${\cal O}(DN)$ calculations, where $D=D_1+D_2+1$.
If $S_j$ is independent of $j$ as $S_j=S$, we can evaluate Eq.\ (\ref{eq:FastViterbiOuterProduct}) slightly more efficiently as
\begin{equation}
\hat{p}_M(i)=b_i(o_M)\max\left\{\max_{j \in {\rm nbh}(i)}[\hat{p}_{M-1}(j)a_{j,i}],
r_iS\max_{j}\left[\hat{p}_{M-1}(j)\right]\right\},
\end{equation}
which is a generalisation of the result in Eq.\ (\ref{eq:FastViterbiUniform}).
Therefore, a fast Viterbi algorithm can be used efficiently for the outer-product HMM if $D\ll N$.
Note that we do not require any pruning to reduce the computational complexity.
This algorithm has slightly larger space complexity than the algorithm described in Sec.\ \ref{sec:UniformModel} because a storage space of $\bm S=(S_i)$ and $\bm r$ is necessary.
The increase in space complexity is only ${\cal O}(2N)$ and does not disrupt real-time processing in practice.

We can also derive efficient algorithms similarly for the forward and backward algorithm of the outer-product HMM, which are important for training HMMs using observation data.
The forward and backward variables are defined as (\cite{Rabiner1989})
\begin{align}
F_m(i)&=P(O_{1:m}=o_{1:m},I_m=i),\\
B_m(i)&=P(O_{m+1:M}=o_{m+1:M}|I_m=i),
\end{align}
and they can be updated with the forward and backward algorithms:
\begin{align}
F_m(i)&=\sum_jF_{m-1}(j)a_{j,i}b_i(o_m),\label{eq:OrgForwardAlg}\\
B_{m-1}(i)&=\sum_ja_{i,j}b_j(o_m)B_m(j).
\end{align}
For the outer-product HMM, we can refine the forward algorithm as
\begin{align}
F_m(i)&=\sum_jF_{m-1}(j)(\alpha_{j,i}+S_jr_i)b_i(o_m)\\
&=\sum_{j \in {\rm nbh}(i)}F_{m-1}(j)\alpha_{j,i}b_i(o_m)+\sum_jF_{m-1}(j)S_jr_ib_i(o_m)\\
&=\sum_{j \in {\rm nbh}(i)}F_{m-1}(j)\alpha_{j,i}b_i(o_m)+r_ib_i(o_m)\left[\sum_jF_{m-1}(j)S_j\right].
\label{eq:FastForwardAlg}
\end{align}
The computational complexity of the original forward algorithm in Eq.\ (\ref{eq:OrgForwardAlg}) is ${\cal O}(N^2)$ since there are $N$ summations over $N$ elements.
In the refined version in Eq.\ (\ref{eq:FastForwardAlg}), the factor in square brackets of the second term is independent of $i$ and as it is sufficient to calculate it once,  the computational complexity is reduced to ${\cal O}(DN)$.
We can derive a refined backward algorithm similarly as
\begin{equation}
B_{m-1}(i)=\sum_{j \in {\rm nbh}(i)}\alpha_{i,j}b_j(o_m)B_m(j)+S_i\left[\sum_jr_jb_j(o_m)B_m(j)\right].
\end{equation}

\subsection{Improvements in score-following results}
\label{sec:DiscOnImprovement}
\subsubsection{Estimation errors in score positions and the following time}\label{sec:EstimationErrorFollowingTime}
We will now discuss how much the results for score following could be improved with the knowledge of musical structures and performer tendencies, which are incorporated in $\bm s=(s_i)$ and $\bm r=(r_i)$ in our model.
There is generally a risk of (i) recognising a performed note in neighbouring chords as a note in a remote chord and of (ii) estimating the wrong score position after skips.
Both of these contribute to errors in score following and the latter is related to the following time of skips.
The risk of the first error occurring generally decreases if $\bm s$ and $\bm r$ are appropriately given, but it is not easy to quantitatively analyse the effect.
On the other hand, we can roughly estimate the dependence for the second error when the probability of performance mistakes is small as shown below.

When a performer resumes after skips, we must observe at least a few notes to identify the score position unless a pitch is associated uniquely with each score position, which is unlikely in practice.
The following time is roughly equal to the time until an almost unique identification of a score position is reached after resumption and the estimation results are uncertain and probably erroneous before the following time.

\subsubsection{Estimating the following time}\label{sec:EstimationOfFT}
We can estimate the following time by using an information-theoretic argument.
First consider that the score is monophonic and is generated by an stochastic process of discrete i.i.d.\ random variables taking $N_p$ values ({\it i.e.}, pitches) with equal probability.
We assume that the length of the score $N$ is sufficiently large.
Given two randomly chosen subsequences of the score of length $L$, the probability that they coincide is $1/N_p^L$.
This means that the number of candidate score positions decreases proportionally to $1/N_p$ for each observed note, and the following time $L_{\rm FT}$ is close to the time when all incorrect candidates are rejected, which is estimated as a sum of the expected time $L_1$ when the number of incorrect candidates becomes unity and the expected time $L_{\rm rej}$ when this candidate is rejected, which are given as $1\simeq N/N_p^{L_1}$ and, $L_{\rm rej}=\displaystyle\sum_{k=1}^\infty k\left[(1/N_p)^{k-1}-(1/N_p)^k\right]=(1-1/N_p)^{-1}$.
Thus,  $L_{\rm FT}\simeq L_1+L_{\rm rej}\simeq{\rm ln}\,N/{\rm ln}\,N_p+(1-1/N_p)^{-1}$ \footnote{Alternatively, we can estimate this directly by calculating the ``decay time'' of all the $N-1\simeq N$ incorrect candidates as
\[  L_{\rm FT}=\sum_{k=1}^\infty k\left[1-(1-1/N_p^{k-1})^N-\left\{1-(1-1/N_p^k)^N\right\}\right]
=1+\sum_{k=1}^\infty\left[1-(1-1/N_p^k) ^N\right]. \]
Although this yields more accurate but close result, we will discuss with the approximate result in the main text since it can be expressed as a simpler explicit form, which clarifies the following discussion.}.
Looking at this more closely, we have a certain chance to obtain a correct score position even if multiple candidates still remain; when two candidates remain, we can identify the correct one with 50\% probability.
This reduces the actual $L_1$ to $L_1\simeq{\rm ln}\,\tilde{N}/{\rm ln}\,N_p
=((\sum_i{\rm ln}\,\tilde{N}_i)/N)/{\rm ln}\,N_p$, where $\tilde{N}_i=\#\{j|j\leq i\}(=i)$
(the choice of order is arbitrary and does not change the result when summed up over all $i$) and $\tilde{N}=\exp((\sum_i{\rm ln}\,\tilde{N}_i)/N)$.
This gives an estimate of the (averaged) following time when we have no prior knowledge on the score positions to where skips occur and if the performance contains no mistakes.
This result can be readily generalised to the case when the score is generated by a Markov process (not necessarily of first order), and the following time is estimated as $L_{\rm FT}\simeq L_1+L_{\rm rej}$ with $L_1\simeq({\rm ln}\,\tilde{N})/h$ and $L_{\rm rej}=(1-e^{-h})^{-1}$, where $h$ is the entropy rate of the Markov process.
This is a consequence of the asymptotic equipartition property (see {\it e.g.}, Ref.~\cite{Cover1991}).
Note that $L_{\rm rej}\simeq 1$ for large $h$, and we can practically approximate $L_{\rm rej}$ with unity.

Next, we consider the case where the distribution $\bm r$, {\it i.e.}, the probability of the resumption positions, is given.
If $\bm r$ is uniformly distributed, the following time is given as above, but if $\bm r$ is not uniform we can more efficiently estimate the score position.
For example, if $\bm r$ is nonzero only for $N'$ score positions ($N'<N$).
A similar argument as above shows that only about $({\rm ln}\,N')/h$ notes are necessary to identify the resumption position if the $N'$ score positions are placed independently of pitch information.
For a general $\bm r$, it can be used as a prior distribution on the score position.
Let us assume that $\bm r$ is distributed independently of pitch information and the correct resumption position is $i$.
Even in this case, the number of candidate score positions decreases proportionally to $e^{-h}$ for each observation, but in this case, when incorrect candidates $j$ with probability $r_j\geq r_i$ are rejected a correct estimate is obtained.
Thus $L_1$ is estimated as $L_1(i)\simeq({\rm ln}\,\tilde{N}_i(\bm r))/h$, where $\tilde{N}_i(\bm r)$ is defined as $\tilde{N}_i(\bm r)=\#\{j|r_j> r_i,$ or $r_j=r_i$ and $j\leq i\}$.
The averaged $L_1$ is then estimated by the expectation value: $L_1\simeq\sum_ir_iL_1(i)\simeq(\sum_ir_i{\rm ln}\,\tilde{N}_i(\bm r))/h$.
Note that this estimate is consistent with the online Viterbi decoding result with the initial probability given by $\bm r$.
The numerator $\sum_ir_i{\rm ln}\,\tilde{N}_i(\bm r)$ is a measure of widespreadness of the distribution $\bm r$ and it is interpreted as the logarithm of the effective number of candidate resumption positions.
Since the nature of the quantity is similar to the entropy, conventionally denoted as $H$, we define $H'(\bm r)=\sum_ir_i{\rm ln}\,\tilde{N}_i(\bm r)$ for later convenience.
Most important of all, the result shows the following time depends logarithmically on the effective number of candidate resumption positions given as the exponential of $H'(\bm r)$.
The above discussion is based on the asymptotic equipartition property, or essentially the law of large numbers, and the above is a good approximation only when ${\rm ln}\,N$ is quite larger than $h$, or $L_{\rm FT}$ is quite larger than unity.

The problem is more complex in several respects in the real performances we are interested in.
Actual scores may have structures like repeated phrases and sections which are not consistent with the Markovian assumption or the assumption of sufficiently large length, and the entropy rate may not be constant over the whole score.
Nevertheless, as long as repetitions are scarce and the distribution $\bm r$ is nearly independent of pitch information, the above result holds approximately for some effective entropy rate $h_{\rm eff}$ since the entropy rate is logarithmically dependent on the number of pitch candidates and its variations are expected to be small.
For polyphonic scores, we can define effective entropy rate $h_{\rm eff}$ of chords if a chord is used as a unit of observation instead of a note.
The dependence of $\bm r$ on the averaged following time is then summarised as
\begin{equation}\label{eq:FTEstimation}
L_{\rm FT}\simeq L_1+L_{\rm rej}\simeq H'(\bm r)/h_{\rm eff}+1.
\end{equation}
If there are performance mistakes, the following time increases on average, and we must stochastically argue the rejection of incorrect candidate resumption positions, which is cumbersome.
However, as long as the probability of mistakes is small, Eq.\ (\ref{eq:FTEstimation}) should hold approximately because the averaged following time should be continuous on the probability of mistakes.
Finally, the effect of $\bm s$ on the following time is limited as long as it is independent of pitch information, since the resumption positions are independent of $\bm s$.

In the above, we have assumed that the distribution $\bm r$ of resumption positions is independent of stop positions as in the outer-product HMM.
If this assumption is not valid, we must return to the full HMM, which is described by the transition probability matrix
\begin{equation}
a_{i,j}=\alpha_{i,j}+g_{i,j},
\end{equation}
where $\alpha_{i,j}$ is same as that in Eq.~(\ref{eq:OuterProductTrMatrix}) and $g_{i,j}$ describes skips, which is generally much smaller than $\alpha_{i,j}$.
Although it is out of scope of this paper to discuss this general case, we can estimate the averaged following time in the general case using the result in Eq.~(\ref{eq:FTEstimation}) as
\begin{equation}
L_{\rm FT}\simeq \sum_iw_iH'(\tilde{\bm g}_i)/h_{\rm eff}+1.
\end{equation}
Here $\tilde{\bm g}_i=(g_{i,j}/\sum_kg_{i,k})_j$ is the normalised distribution of resumption positions of skips from stop position $i$, and $w_i$ is the probability that a skip is from stop position $i$, normalised as $\sum_iw_i=1$.

The number of estimation errors $E$ of the second kind described in Sec.\ \ref{sec:EstimationErrorFollowingTime} is roughly estimated as a function of the averaged following time $L_{\rm FT}$.
If $N_{ch}$ is the averaged number of pitches contained in a chord, the averaged number of estimation errors per skip $E$ is given as $E=N_{ch}L_{\rm FT}$, assuming that the distributions of the resumption positions are independent of the pitch content of chords.
Thus, if the averaged following time is reduced by $\Delta L_{\rm FT}$, we can expect a reduced number of estimation errors estimated as $\Delta E=N_{ch}\Delta L_{\rm FT}$.

\subsubsection{Simulation results for dependence of the following time on $H'({\bf r})$}
\begin{figure}[tbp]
\begin{center}
\subfigure[Results for Debussy's ``La fille aux cheveux de lin.'']
{\includegraphics[clip,width=0.5\columnwidth]{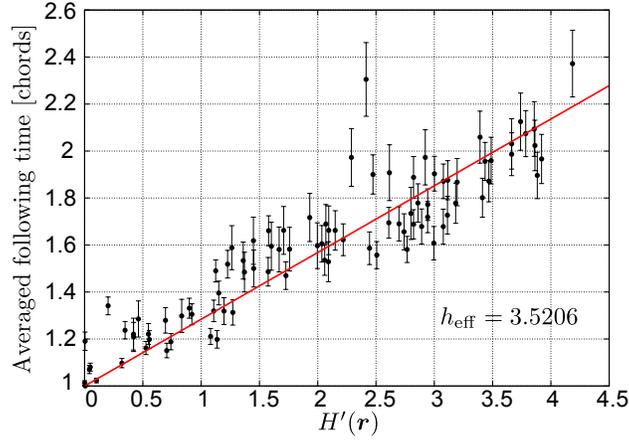}}\\
\subfigure[Results for Mozart's sonata for two pianos.]
{\includegraphics[clip,width=0.5\columnwidth]{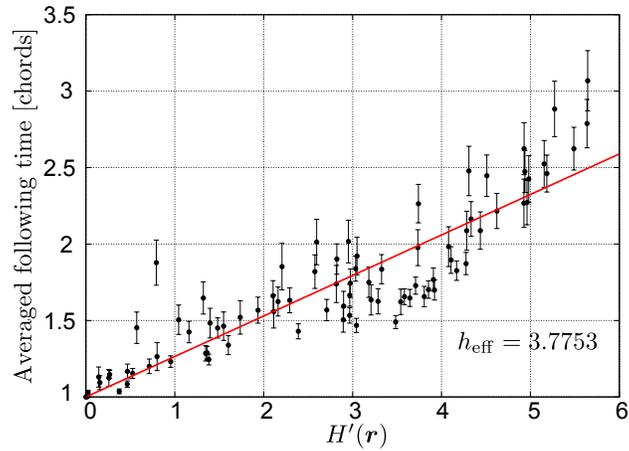}}\\
\subfigure[Results for Mussorgsky's Promenade.]
{\includegraphics[clip,width=0.5\columnwidth]{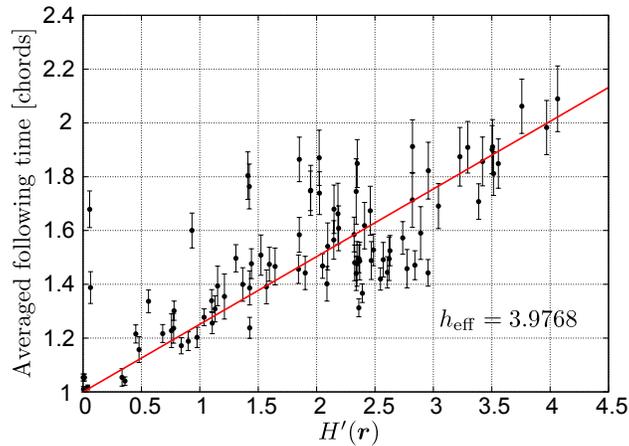}}
\end{center}
\caption{Simulation results for dependence of the averaged following time on the distribution $\bm r$. Each point corresponds to a randomly generated distribution $\bm r$, and the averaged following time measured by score following for synthetic performance is indicated with statistical errors. Fitted value of $h_{\rm eff}$ is also shown.}
\label{fig:SimulationFT}
\end{figure}
We conducted a computer simulation to confirm the above result and measured the following time for synthetic performances with various distributions of resumption positions.
For the three polyphonic pieces described in Secs.\ \ref{sec:RepeatSkip} and \ref{sec:PerformancePreparation}, we first randomly generated distributions $\bm r$ with various $H'(\bm r)$ and then generated synthetic performances with skips created stochastically with the distribution $\bm r$.
We generated 300 skips for each synthetic performance and measured and averaged the following time, which is defined as the number of chords played after resumption until a correct estimation is obtained, with the score-following algorithm using the outer-product HMM with the same distribution $\bm r$ used for synthesising the performances.
Performance mistakes were not included in the synthetic performance, but they were included in the HMM used for score following.
The distribution $\bm s$ was set to the uniform distribution.

The results in Fig.\ \ref{fig:SimulationFT} illustrate the overall linear dependence of the averaged following time on $H'(\bm r)$, which confirms the result above.
A fitted linear line is plotted with the value of $h_{\rm eff}$.
We find some points are far above the fitted line, and we checked that they are influenced by repeated phrases in the scores.
Although a precise prediction is difficult, Eq.\ (\ref{eq:FTEstimation}) yields a rough estimate.
Interestingly, we see $h_{\rm eff}\approx 3.5$--4 for all of the three pieces.

When $\bm r$ is distributed uniformly, $H'(\bm r)=(\sum_{i=1}^N{\rm ln}\,i)/N\equiv H'(N)$, and $H'(N)\approx {\rm ln}\,N-1$ for $N\gg1$.
For instance, $N=249$ and $H'(N)\simeq 4.5$ for the Debussy's piece in Fig.\ \ref{fig:SRPositionsDeb}, and $H'(\bm r)$ for the distribution in the figure is $H'(\bm r)\simeq 2.0$.
Similarly, $H'(N)\simeq 5.7$ (resp.~4.2) and $H'(\bm r)\simeq 2.2$ (resp.~1.4) for the Mozart's (resp.~Mussorgsky's) piece in Fig.\ \ref{fig:SRPositionsMoz} (resp.~Fig.~\ref{fig:SRPositionsMus}).
Thus, a reduction factor of $H'(\bm r)/H'(N)\approx 0.44$ (resp.~$0.39$, $0.33$) is expected for the averaged following time (minus unity) for the Debussy's (resp.~Mozart's, Mussorgsky's) piece, which also contributes to reducing some fraction of the estimation errors.

\section{Evaluation of the score-following algorithms}\label{sec:Evaluation}
\subsection{Quantitative evaluation of computational complexity}\label{sec:EvalCompComplex}
We carried out a computer simulation to measure the processing time to quantitatively evaluate the reduction in computational complexity with the fast Viterbi algorithm described in Sec.\ \ref{sec:InferenceAlg}.
Because the processing time is mainly dependent on $D$ and $N$ (the number of chords in the performance score) and virtually independent of the details of score content and observation events, we used synthetic scores with various lengths from $N=100$ to $N=21500$ and a random pitch sequence with a length of 100 for the observation events.
The computational environment involved an Intel(R) Core(TM) i5-2540M CPU, with 8 GB of RAM and the Windows 7 64-bit Professional OS.

\begin{figure}[tb]\begin{center}
\includegraphics[clip,width=0.7\columnwidth]{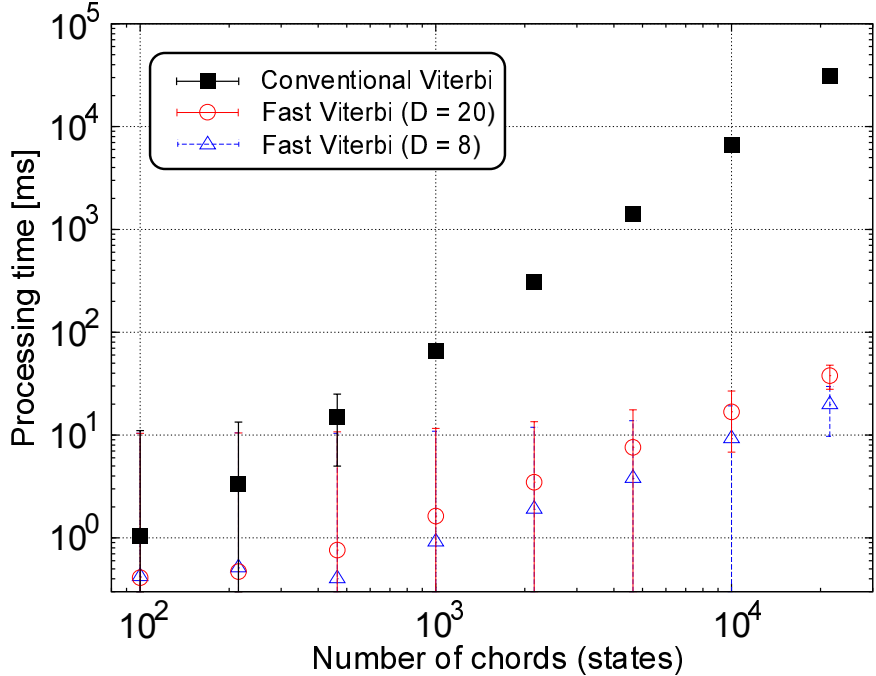}
\caption{Averaged processing time for Viterbi update. The squares indicate results using the online version of conventional Viterbi algorithm and the triangles and circles indicate results using the fast Viterbi algorithm in Eq.\ (\ref{eq:FastViterbiOuterProduct}), for $D=8$ and 20. Errors indicate $1\sigma$ confidence intervals including statistical and systematic errors (see text).}
\label{fig:ProcessingTime}
\end{center}
\end{figure}
The results of the measurement are shown in Fig.\ \ref{fig:ProcessingTime}, where the squares indicate the averaged processing time using the online version of conventional Viterbi algorithm in Eq.\ (\ref{eq:ViterbiUpdate}) and the triangles and circles indicate the results using the fast Viterbi algorithm in Eq.\ (\ref{eq:FastViterbiOuterProduct}), for $D=8$ and 20.
The error includes both the statistical error and the systematic error in the computer system, which was taken as 10 ms.
The results show that the increase in processing time with the number of chords is significantly suppressed with the fast Viterbi algorithm.
Especially, the processing time increases significantly for $N\gtrsim 1000$ with the conventional Viterbi algorithm while it stays within a few tens of milliseconds up to about 10000 chords with the fast Viterbi algorithm.
Although not shown in the figure, we also measured the processing time of the fast Viterbi algorithm for the uniform skip model in Sec.\ \ref{sec:UniformModel}, and confirmed that it is slightly less than that for the outer-product model.

An acceptable delay time for score following depends on particular applications.
One of the most severe situations is with automatic accompaniment, where a delay of a few hundred milliseconds is serious \cite{Cont2007}.
An effective way of doing accompaniments, particularly in situations involving musical practice, is to play accompaniments so that they instantaneously correspond to observed events, and in this most severe case, the delay must be suppressed within a few tens of milliseconds \cite{Nakamura2013}.
Since there are also delays resulting from signal input, signal output and the accompaniment algorithm, an acceptable processing time of score following is then about $10$ ms.
This corresponds to an upper limit of ${\cal O}(100)$ chords for the conventional Viterbi algorithm, which is the size of short musical pieces.
On the other hand, the fast Viterbi algorithm with smaller $D$ can cover typical concert-size pieces containing ${\cal O}(10^3$--$10^4)$ chords.
The processing time is almost proportional to $D$, while the descriptive power of the model increases for larger $D$.
The result indicates that for a score with 10000 chords, $D\approx 10$ is a practical upper bound for $D$, and larger $D$ can be used for smaller $N$.
Finally, we need to comment that the absolute value of the result is dependent on the processing power of computers, but the relative value remains (almost) unchanged and the reduced computational complexity attained by the fast Viterbi algorithm always remains effective.

\subsection{Preparation of performance data}\label{sec:PerformancePreparation}

For the purpose of analysing human performances to set model parameters and evaluating the score-following algorithms, we collected performances by three pianists (two amateurs and one professional) who played the same pieces.
The musical pieces used are Debussy's ``La fille aux cheveux de lin'' (No.\ 8 of Pr\'eludes Premier Livre), Mozart's sonata for two pianos in D major K.\ 448 (the first piano part of the exposition in the first movement) and Mussorgsky's ``Promenade'' (the first piece from the suite ``Pictures at an Exhibition'').
For the ease of collection and analysis of data, relatively shorter pieces were selected:
The numbers of chords/onsets in the scores were 249/583, 771/1381 and 181/675.
Since our focus is on performances during practices, the pianists were advised to practice the pieces freely.
The pianists played on a digital piano and the performances were recorded as MIDI data.
The length of each practice varied between 15 and 45 min.
We also recorded play-through performances after practice to evaluate the score-following algorithms in situations with few skips but with performance mistakes.

The technical level and familiarity of the three pianists with the pieces varied.
None of them were very familiar with the pieces.
The first and amateur performer (A) had played all of the three pieces, but had not practiced them many times.
The second and amateur performer (B) had practiced the Debussy's and Mussorgsky's pieces before, and practiced the Mozart's piece for the first time.
The third and professional performer (C) had played the Debussy's piece and the second piano part of
Mozart's sonata, but had no experience with playing the first part of the sonata or the Mussorgsky's piece.

\begin{table}[tb]
\caption{Number of mistakes and skips in the play-through performances. Here ``Insertion'' and ``Deletion'' means chord insertion and deletion errors, and ``Skip'' counts the number of repeats/skips involving more than three skipped chords.}
\label{tab:MistakesRareSkip}
\begin{center}
\begin{tabular}{c|c|cccc}\hline\hline
Piece (Performer) &Onset&Pitch error&Insertion&Deletion&Skip\\
\hline
Debussy (A) &1210&34&9&3&1\\
Debussy (B) &668&73&22&3&1\\
Debussy (C) &1159&28&15&2&0\\
\hline
Mozart (A) &2770&100&23&20&0\\
Mozart (B) &1540&140&25&8&1\\
Mozart (C) &1372&15&0&2&0\\
\hline
Mussorgsky (A) &2066&86&2&0&0\\
Mussorgsky (B) &708&67&3&0&0\\
Mussorgsky (C) &2093&97&2&1&0\\
\hline\hline
\end{tabular}
\end{center}
\end{table}
\begin{table}[tbp]
\caption{Number of mistakes and skips in the sampled performances during practice. Ten performance segments each containing 500 note onsets were randomly chosen as samples for each piece. The mistakes have the same meanings as those in Table \ref{tab:MistakesRareSkip}.}
\label{tab:MistakesWithSkip}
\begin{center}
\begin{tabular}{cc|cccc}\hline\hline
Piece & No. (Performer) &Pitch error&Insertion&Deletion&Skip\\
\hline
Debussy&1 (C)&17&19&5&7\\
&2 (A)&11&4&1&6\\
&3 (C)&7&7&2&9\\
&4 (C)&17&25&0&8\\
&5 (C)&12&6&0&6\\
&6 (A)&11&2&0&5\\
&7 (C)&37&14&0&4\\
&8 (A)&4&1&0&4\\
&9 (B)&58&60&7&12\\
&10 (B)&41&39&3&12\\
\hline
Debussy&Total&215&177&18&73\\
\hline\hline
Mozart&1 (C)&9&4&2&11\\
&2 (A)&26&0&3&2\\
&3 (A)&7&3&1&6\\
&4 (B)&56&21&5&12\\
&5 (B)&49&24&3&7\\
&6 (A)&17&1&5&4\\
&7 (C)&27&20&0&19\\
&8 (B)&54&11&2&7\\
&9 (C)&5&3&0&8\\
&10 (A)&11&19&12&17\\
\hline
Mozart&Total&261&106&33&93\\
\hline\hline
Mussorgsky&1 (B)&23&13&0&2\\
&2 (C)&29&3&0&2\\
&3 (B)&42&21&0&3\\
&4 (A)&15&0&0&1\\
&5 (B)&35&24&0&1\\
&6 (C)&21&6&0&5\\
&7 (A)&24&1&0&2\\
&8 (C)&34&9&0&1\\
&9 (C)&19&5&0&3\\
&10 (C)&29&3&0&1\\
\hline
Mussorgsky&Total&271&85&0&21\\
\hline\hline
\end{tabular}
\end{center}
\end{table}
The recorded performances were analysed by the authors and the performed notes were matched to the notes in the scores.
Most of the notes were unambiguously matched, but there were notes that were difficult to associate with any notes in the scores.
They typically appeared at skips when the performers looked for the score position to resume playing by touching the keyboard halfheartedly, and perhaps unconsciously.
While these ``non-associated'' notes were not used in the analyses, they were included in evaluations since they naturally appear in real situations.
Because the recorded performances during practice were long, we divided them into segments of 500 notes and we randomly chose ten segments for precise analysis and evaluation.
In Tables \ref{tab:MistakesRareSkip} and \ref{tab:MistakesWithSkip}, the numbers of performance mistakes and skips are listed for the play-through performances and the performances during practice.
A repeatedly performed chord with a chord IOI longer than 35 ms or a transition to the previous three chords is counted as a chord insertion, a transition skipping one or two chords is counted as a chord deletion, and larger repeats or skips are counted as skips.

To analyse large skips and evaluate the following time, we annotated score positions around large skips for all of the performances during practice.
We used all skips with $j-i\leq -4$ or $j-i\geq 4$ ($i$ and $j$ denote the stop and resumption positions) for this, and other smaller skips were treated as chord insertions and deletions.
The number of skips were 288 (resp.~373, 83) for the Debussy's (resp.~Mozart's, Mussorgsky's) piece.

\subsection{Model parameters obtained from performance data}\label{sec:ModelParameters}

\begin{table}[tb]
\caption{List of values for the transition probability obtained from performance data.}
\label{tab:TrProbValues}
\begin{center}
\begin{tabular}{c|c||c|c||c|c||c|c}\hline\hline
$\delta i$&$a_{\delta i}$&$\delta i$&$a_{\delta i}$&$\delta i$&$a_{\delta i}$&$\delta i$&$a_{\delta i}$\\
\hline
$-17$&0.00015&$-11$&0.00022&$-5$&0.00153&1&0.84531\\
$-16$&0.00044&$-10$&0.00051&$-4$&0.00218&2&0.00610\\
$-15$&0.00058&$-9$&0.00065&$-3$&0.00509&3&0.00073\\
$-14$&0.00044&$-8$&0.00124&$-2$&0.00516&4&0.00029\\
$-13$&0.00029&$-7$&0.00182&$-1$&0.00886&5&0.00015\\
$-12$&0.00007&$-6$&0.00073&0&0.11342&6&0\\
\hline\hline
\end{tabular}
\end{center}
\end{table}
The parameters of the performance HMM can be set using the performance data.
First, let us discuss the transition probability.
Although the transition probability $a_{i,j}$ can vary at each score position $i$, it is difficult to obtain them all independently for the lack of huge size of performance data.
We therefore gather relative information on $j-i$ and use the averaged value for all $i$, {\it i.e.}, we set $a_{i,j}=a_{j-i}$.
The transition probabilities $a_{\delta i}$ for $-17\leq \delta i\leq 6$ obtained from the performance data are listed in Table \ref{tab:TrProbValues}.
The $a_{\delta i}$ for $\delta i$ out of the range, which corresponds to large skips, was very small (see Fig.\ \ref{fig:diffSRPositions}).

\begin{table}[tb]
\caption{List of referential values of $D$, $D_1$, $D_2$ and $\bar{\gamma}$.}
\label{tab:ListD1D2}
\begin{center}
\begin{tabular}{c|c|c|c}\hline\hline
$D$ & $D_1$ & $D_2$ & $\bar{\gamma}$\\ \hline
$4$ & $1$ & $2$ & $0.02630$\\
$10$ & $7$ & $2$ & $0.00981$\\
$20$ & $15$ & $4$ & $0.00480$\\
\hline\hline
\end{tabular}
\end{center}
\end{table}
The parameters, $D_1$ and $D_2$, must be fixed to certain values, which define the band matrix $\alpha_{i,j}=a_{j-i}$ for $-D_1\leq j-i\leq D_2$, and the transitions with $i$ and $j$ out of this range are described by $\bar{\gamma}$ or $\gamma=\bar{\gamma}/N$.
Although there are no strict principles to choose $D_1$ and $D_2$, a reasonable way is to choose them by maximising $\sum_{-D_1\leq\delta i\leq D_2}a_{\delta i}$ for a fixed $D=D_1+D_2+1$.
We list three sets of these values chosen in this manner and corresponding $\bar{\gamma}$ for $D=4$, 10 and 20 in Table \ref{tab:ListD1D2} for later reference.

The distribution of the stop positions $\bm s$ and of the resumption positions $\bm r$ can also be extracted from the performance data.
The distributions for the three pieces can be obtained by normalising the histograms shown in Figs.\ \ref{fig:SRPositionsDeb}, \ref{fig:SRPositionsMoz} and \ref{fig:SRPositionsMus}.
Since the data are rather sparse, we prepared for each performer the distribution obtained from performances of the other two performers, which is used in evaluations with cross validation, in addition to the distribution of all three performers.
To cure the zero frequency problem, we uniformly added a constant 0.01 to each bin corresponding to score positions before normalising the histograms in each case.

\begin{table}[t]
\caption{List of values for the output probability obtained from the performance data. The ``chord'', ``s.t.'', ``w.t.'', ``oct'' and ``rest'' indicate probabilities of correct pitches, mistakes by semitones, whole-tones, octaves, and other mistakes (see text).}
\label{tab:OutProbValues}
\begin{center}
\begin{tabular}{c|c|c|c|c|c}\hline\hline
$A$ & chord & s.t. & w.t. & oct & rest\\ \hline
$p_A$ & 0.9497 & 0.0145 & 0.0224 & 0.0047 & 0.0086\\
\hline\hline
\end{tabular}
\end{center}
\end{table}
The values of the output probability can also be determined from the performance data, but it is again necessary to tie up parameters to avoid the problem of data sparseness.
Since pitch errors by semitones, whole-tones and octaves are common, we use the following empirical parametrisation
\begin{equation}
P(O_m=o|I_m=i)=p_A/|c^A_i|\quad\text{if $o\in c^A_i$}\quad
(A={\rm chord},{\rm s.t.},{\rm w.t.},{\rm oct},{\rm rest}).
\end{equation}
Here $c^{\rm chord}_i=c_i$ is the set of correct pitches in the chord, $c^{\rm s.t.}_i$ (resp.~$c^{\rm w.t.}_i$, $c^{\rm oct}_i$) is the set of erroneous pitches by semitones (resp.~whole-tones, octaves), exclusively defined as $c^{\rm s.t.}_i=\{o\pm1|o\in c_i\}\backslash c_i$ (resp.~$c^{\rm w.t.}_i=\{o\pm2|o\in c_i\}\backslash (c_i\cup c^{\rm s.t.}_i)$,
$c^{\rm oct}_i=\{o\pm12|o\in c_i\}\backslash (c_i\cup c^{\rm s.t.}_i\cup c^{\rm w.t.}_i)$), $c^{\rm rest}_i$ is the set of residual pitches, and the symbol $|\cdot|$ denotes the number of pitches contained in each set.
The $p_A$ is the probability that $o$ is in the set $c_i^A$, normalised as $\sum_Ap_A=1$, and is parametrised independently of $i$.
The values of $p_A$ obtained from the performance data is listed in Table \ref{tab:OutProbValues}.

\subsection{Evaluation of estimations of score positions}\label{sec:ScorePositionEstEvaluation}
In this section, we describe the results of our score-following algorithms for the human performances.
The purposes are to evaluate overall quality of the algorithms and discuss improvements with the outer-product HMM for real performances with mistakes since the theoretical discussion in Sec.~\ref{sec:DiscOnImprovement} assumed no or rare mistakes.

\subsubsection{Results for the play-through performances}
To evaluate the score-following algorithms, we implemented algorithms using the outer-product HMM, the uniform skip model, and a model without modeling large skips, which is obtained by setting $\gamma=0$ in the uniform skip model.
These algorithms are denoted by O (for outer-product), U (for uniform) and N (for no skips).
The algorithm N only treats transitions within neighbouring chords and it is essentially the same as the one proposed in Ref.~\cite{Bloch1985}.
For the algorithm O, we used the distributions $\bm s$ and $\bm r$ obtained from the performance data by all the performers (see Sec.~\ref{sec:ModelParameters}).

\begin{table}[tb]
\caption{Error rate (\%) for the play-through performances of the online and offline algorithms. Algorithm O, U and N denote those using the outer-product HMM, the uniform skip model and the model without modeling large skips, and online and offline are abbreviated as ``on'' and ``off''. We take $D=10$ for three algorithms.}
\label{tab:AccRareSkip}
\begin{center}
\begin{tabular}{c|c|c|c}\hline\hline
Algorithm & Debussy & Mozart & Mussorgsky\\
\hline
O-on&3.24&5.03&1.87\\
U-on&2.78&5.00&1.93\\
N-on&2.05&5.00&1.71\\
\hline
O-off&0.53&3.04&0.62\\
U-off&0.46&2.83&0.76\\
N-off&0.46&3.04&0.60\\
\hline\hline
\end{tabular}
\end{center}
\end{table}
We first present the results of evaluation for the play-through performances.
Since skips are rare in these performances, we only evaluated the error rate of score following, which is the proportion of mismatched notes to the total number of performed notes.
The non-associated notes were not used in calculating the error rate.
The results are summarised in Table \ref{tab:AccRareSkip}, where we set $D=10$ for all three algorithms.
A comparison of the results with different $D$ is given in the Appendix.
The online algorithms can be modified to offline algorithms by implementing the back-tracking of the Viterbi updates.
The error rates for the offline algorithms are also listed in the table.

The error rate for the algorithm N was the smallest in most cases, as expected from the fact that the performances only contain a few skips.
We see that the error rates for the online algorithms O and U are only a few percents more than the results for the algorithm N, indicating that the modeling of arbitrary skips does not greatly increase the error rate of score following for performances with many mistakes but rare skips.
All the error rates for the online algorithms are under about $5\%$ and thus the algorithms are shown to be robust against many performance mistakes.
The error rates for the offline algorithms are always lower than those for the online algorithms, illustrating the fact that the use of the future information improves the matching accuracy.
A major contribution to the estimation errors is made by arpeggios and arpeggiated chords in the Debussy's piece, and trills and short appoggiaturas in the Mozart's piece.
Another contribution is due to misidentification of inserted notes/chords.
In the Mozart's piece, there were occasions where the synchronicity of both hands was weakened, particularly in fast passages, and these were also a source of estimation errors.

\subsubsection{Results for the performances during practice}
\begin{table}[tb]
\caption{Error rate (\%) and its statistical error for the performances during practice of the online and offline algorithms. Same abbreviations for the algorithms are used as in Table \ref{tab:AccRareSkip} and we again take $D=10$ for all algorithms. For the distributions $\bm s$ and $\bm r $, that obtained from the manually labeled data leaving each performer is indicated as ``CV'', that obtained from the estimation results of the offline algorithm using the uniform skip model is indicated as $\bm s,\bm r|_{\rm offline}$.}
\label{tab:AccWithSkip}
\begin{center}
\begin{tabular}{c|c|c|c}\hline\hline
Algorithm & Debussy & Mozart & Mussorgsky\\
\hline
O-on &$10.1\pm2.2$&$9.3\pm1.8$&$8.7\pm1.4$\\
O-on (CV) &$11.7\pm2.4$&$10.8\pm2.0$&$8.0\pm1.0$\\
O-on($\bm s,\bm r|_{\rm offline}$) &$10.4\pm2.1$&$10.0\pm1.8$&$7.3\pm1.0$\\
O-on($\bm s,\bm r|_{\rm offline}$) (CV) &$11.7\pm2.2$&$11.4\pm1.9$&$8.1\pm0.9$\\
U-on &$13.9\pm2.3$&$12.1\pm1.7$&$8.3\pm1.2$\\
N-on  &$21.3\pm3.1$&$23.1\pm3.6$&$8.3\pm1.1$\\
\hline
O-off &$4.3\pm1.6$&$4.6\pm1.1$&$1.9\pm0.5$\\
O-off (CV) & $4.3\pm1.6$ &$4.8\pm0.9$&$2.0\pm0.5$\\
O-off($\bm s,\bm r|_{\rm offline}$) &$5.3\pm1.7$&$4.8\pm1.1$&$2.1\pm0.5$\\
O-off($\bm s,\bm r|_{\rm offline}$) (CV) &$5.4\pm1.7$&$4.9\pm0.9$&$2.2\pm0.5$\\
U-off &$5.2\pm1.5$&$4.9\pm1.1$&$2.1\pm0.5$\\
N-off &$13.7\pm2.8$&$11.9\pm2.0$&$3.7\pm0.9$\\
\hline\hline
\end{tabular}
\end{center}
\end{table}
We next present the results for the performances during practice.
The error rates for the algorithms are listed in Table \ref{tab:AccWithSkip}, where the averaged error rate over the performance segments described in Sec.\ \ref{sec:PerformancePreparation}, together with the $1\sigma$ statistical error, is shown.
We used the same algorithms as in the previous section.
On average, the algorithm O outperforms the algorithm U, which greatly outperforms the algorithm N, showing that the explicit modeling of arbitrary skips and the use of the information on the stop and resumption positions are effective.
For the Mussorgsky's piece, the error rate for the algorithm O-on is greater than that for the algorithms U-on and N-on, and the difference is less than the $1\sigma$ statistical error.
The main reason that the error rate is not reduced with the algorithm O in the Mussorgsky's piece is that there were a few large skips in the performance samples. Out of 21 skips with $|\delta i|>3$, with $\delta i$ denoting the difference between the resumption and stop positions, only two are those with $|\delta i|>10$ and the largest one was $\delta i=-18$.

Since the data on stop and resumption positions were sparse, we also performed leave-one-out cross-validation: when following performer A, the distributions obtained from the performances B and C are used, and so on.
The results are indicated with ``CV'' in Table \ref{tab:AccWithSkip}.
The results for O-on (CV) is better than U-on, but worse than O-on.
This is as expected from the sparseness of data on stop and resumption positions.

In Table \ref{tab:AccWithSkip}, we also list the results for the algorithm O-on (resp.~O-off) with the distributions $\bm s$ and $\bm r$ obtained from the estimation results of score positions using the algorithm U-off, which we refer to as O-on($\bm s,\bm r|_{\rm offline}$) (resp.~O-off($\bm s,\bm r|_{\rm offline}$)).
For the Debussy's and Mozart's pieces, the error rates for the algorithm O-on($\bm s,\bm r|_{\rm offline}$) are slightly higher than those for the algorithm O-on and are lower than those for the algorithm U-on.
For the Mussorgsky's piece, the error rate is lower than that for the algorithm O-on, but again the differences are near the $1\sigma$ value.
Similar results using the estimation results of the algorithm U-off, but with cross validation using the data without the performer to follow, is also shown in Table \ref{tab:AccWithSkip} (indicated with O-on($\bm s,\bm r|_{\rm offline}$) (CV)).
We see the values are near or slightly larger than those of O-on (CV).

The error rate of the offline algorithms are significantly lower than that of the online algorithms.
This is because the future information reduces ambiguities in estimating the correct score position among other possible candidates soon after skips and recognising insertion and deletion errors, which is one of the main causes of estimation errors by the online algorithms.

\begin{table}[tb]
\caption{Following rate (FR), averaged following time (FT), together with its statistical error, and its standard deviation (SD) of the online algorithms for the performances during practice.
The algorithms and conditions are the same as in Table \ref{tab:AccWithSkip}.}
\label{tab:FTWithSkip}
\begin{center}
\begin{tabular}{c|ccc}\hline\hline
\multicolumn{1}{c|}{Algorithm}&\multicolumn{3}{c}{Debussy}\\\cline{2-4}
 & FR (\%) & FT [chord] & SD [chord] \\
 \hline
O-on & 93.7 & $2.15\pm0.16$ & $2.55$\\
O-on (CV) &93.7&$2.41\pm0.17$&$2.74$\\
O-on($\bm s,\bm r|_{\rm offline}$) & 95.2 & $2.27\pm0.16$ & $2.59$\\
O-on($\bm s,\bm r|_{\rm offline}$) (CV) &94.8&$2.51\pm0.18$&$2.79$\\
U-on & 93.7 & $3.25\pm0.24$ & $3.79$\\
N-on & (73.8) & $\geq 6.29\pm0.49$ & ($7.70$) \\
\hline\hline
\multicolumn{1}{c|}{Algorithm}&\multicolumn{3}{c}{Mozart}\\\cline{2-4}
 & FR (\%) & FT [chord] & SD [chord] \\
 \hline
O-on & 98.7 & $2.24\pm0.10$ & $1.99$\\
O-on (CV) &97.6&$3.09\pm0.15$&$2.89$\\
O-on($\bm s,\bm r|_{\rm offline}$) & 98.4 & $2.44\pm0.11$ & $2.22$\\
O-on($\bm s,\bm r|_{\rm offline}$) (CV) &98.1&$3.12\pm0.14$&$2.74$\\
U-on & 97.9 & $3.30\pm0.14$ & $2.66$\\
N-on & (85.5) & $\geq 7.73\pm0.65$ & ($12.55$) \\
\hline\hline
\multicolumn{1}{c|}{Algorithm}&\multicolumn{3}{c}{Mussorgsky}\\\cline{2-4}
 & FR (\%) & FT [chord] & SD [chord] \\
 \hline
O-on & 100 & $1.80\pm0.16$ & $1.50$\\
O-on (CV) &100&$1.78\pm0.15$&$1.37$\\
O-on($\bm s,\bm r|_{\rm offline}$) & 98.8 & $1.80\pm0.19$ & $1.69$\\
O-on($\bm s,\bm r|_{\rm offline}$) (CV) &100&$1.84\pm0.17$&$1.55$\\
U-on & 96.4 & $1.95\pm0.19$ & $1.74$\\
N-on & (69.9) & $\geq 4.98\pm0.56$ & ($5.06$) \\
\hline\hline
\end{tabular}
\end{center}
\end{table}
In evaluating online algorithms, how much and how fast the algorithms can follow skips are also important.
To measure the latter, we can use the following time, which is now defined as the number of chords necessary to achieve a correct estimate of two successive chords after skips.
The reason for the additional condition for correctly estimated successive chords is to ensure the correct matching is not a consequence of randomness.
If a skip is succeeded by another skip before a score position is correctly estimated, then we define the following time to be the number of chords between the two stop positions.
The following rate is defined by the rate that the algorithm correctly estimates score positions before the next skip.

In Table \ref{tab:FTWithSkip}, the averaged following time and following rate are listed for the online algorithms.
We see that the averaged following time is improved with the algorithms using the outer-product HMM compared to the algorithm using the uniform skip model.
The results of cross validation are again better than the uniform skip model, but worse than the results with the closed data.
The algorithm without modeling skips has a low following rate and much longer averaged following time.
Since the algorithm could not correctly estimate score positions in regions annotated by humans in some cases, these values are only approximated values and the displayed value of the averaged following time is its lower bound.
However, this is sufficient to confirm the necessity for modeling skips in score following of performances in practical situations.
A comparison of the result with different $D$ is given in the Appendix.

\subsection{Discussion of the evaluation results}
\label{sec:DiscEvalutionResult}
The evaluation results in the previous section show that modeling skips is indeed significant for score following of practical performances with skips and mistakes.
The algorithm proposed in Ref.~\cite{Bloch1985} can only follow skips within the neighbouring chords.
The range of the neighbouring chords can be widened by setting larger $D$, but as we have seen in Sec.\ \ref{sec:EvalCompComplex}, there is a practical upper bound for real-time working, $D\lesssim 10$.
A similar situation occurs if we simply generalise the algorithms in Refs.~\cite{Tekin2004,Pardo2005,Oshima2005}, since the computational complexity increases if we increase the number of possible resumption positions.
The score-following algorithms using the outer-product HMM and the uniform skip model can handle arbitrary skips, without serious increases in computational complexity.

The score-following algorithm using the outer-product HMM has a lower error rate and a shorter averaged following time than the algorithm using the uniform skip model proposed in Ref.~\cite{Nakamura2013}, as is expected in Sec.\ \ref{sec:DiscOnImprovement}.
As discussed in the section, the degree of improvement depends on the performance score and the tendencies of the performance, particularly the distributions $\bm s$ and $\bm r$.
The reduction factors for the averaged following time (minus one) are $0.51$ for the Debussy's piece and $0.54$ for the Mozart's piece, and they are slightly larger, which is partly due to mistakes in the performances, but roughly in agreement with the estimated values in Sec.\ \ref{sec:DiscOnImprovement}.
The decrease in the error rate resulting from the reduction of the averaged following time is estimated to be $\Delta E\simeq 3.8\%$ and $3.5\%$, with $N_{ch}\simeq 2.34$ and $1.79$ ($N_{ch}$ denotes the averaged number of pitches contained in a chord, see Sec.~\ref{sec:EstimationOfFT}), which is consistent with the results in Table \ref{tab:AccWithSkip}, suggesting that the reduction of the following time is the main cause for the reduction of estimation errors.
The rates of improvements depend on musical pieces and tendencies of performances in general, and the discussion in Sec.~\ref{sec:DiscOnImprovement} yields theoretical estimation for the general case.

The results of cross validation confirm the importance of obtaining appropriate parameters for $\bm s$ and $\bm r$.
In general, results using closed data and that of cross validation should be close when sufficient amount of data are given, and the discrepancies seen in our evaluations are explained by relatively small size of data and/or gaps between tendencies of different performers. In fact, we have confirmed that the larger error rate and following time seen in the Debussy's and Mozart's pieces were mainly due to many repeats of specific phrases by individual performers that are not so frequently seen for the other performers. 
When these tendencies are appropriately obtained, for example, by using past performance data by the same performer, the results of score following would be significantly improved as we calculated in Sec.~\ref{sec:DiscOnImprovement}, when wrong data are used, on the other hand, the improvements would be limited.
To study how these tendencies differ between different performers or situations and how they can be efficiently obtained is of importance and left for further investigations.

The accuracy of estimating score positions with the online version of the algorithm is low compared to the reported values for other offline score-performance matching algorithms, studied for example in Refs.~\cite{Heijink2000,Gingras2011}, even for performances with rare skips.
The main reason is probably that the performances used in the evaluations in this paper are more practical with more performance mistakes, but further studies need to be done to compare the algorithms in detail.
Since the algorithms proposed in these references do not handle skips, the offline versions of the algorithms proposed in this paper can be effective for performances with skips.

The distributions on the stop and resumption positions can be obtained by analysing performance data as we did in this paper, but it is hard in practice to manually analyse data for many musical pieces.
We have shown that the use of estimated results with the offline algorithm using the uniform skip model provides quite improved results, which are almost as good as those with the manually analysed data.
This suggests the possibility of automatically obtaining the distributions with the offline algorithm as long as a sufficient amount of performance data is given:
If performance data by various performers are given, it is possible to obtain distributions which fit to generic performers.
Or if a certain amount of rehearsal data by a specific performer is given, it can be used to improve score following for later rehearsals by the same performer.

\begin{figure}[tbp]
\begin{center}
\includegraphics[clip,width=0.75\columnwidth]{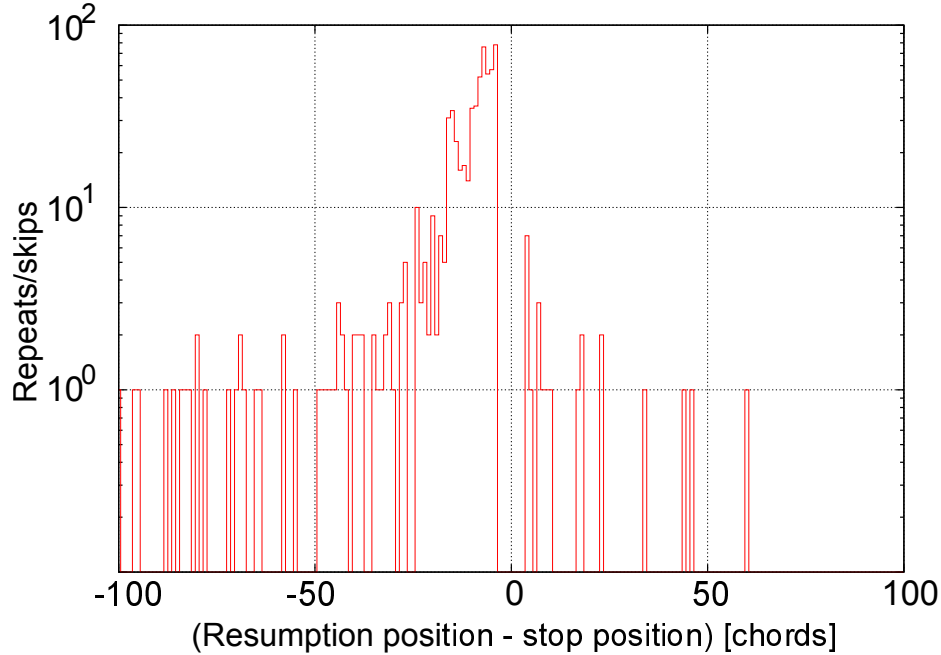}
\end{center}
\caption{Distribution of the differences between the stop and resumption positions in the performances during practice. The data is extracted from performances of Debussy's, Mozart's and Mussorgsky's piece described in Sec.\ \ref{sec:PerformancePreparation}.}
\label{fig:diffSRPositions}
\end{figure}
In the outer-product HMM, we assumed that the distributions of the resumption positions are independent of the stop positions.
In analysing the performance data, we noticed that there is a tendency for forward skips to be less frequent than backward skips (repeats), and skips to more distant score positions to be rarer than those to nearer score positions.
The distribution of the differences between the stop and resumption positions $\delta i$ in the range $-100\leq \delta i\leq 100$ is shown in Fig.\ \ref{fig:diffSRPositions} for all performance data from the three pieces during practice.
Skips out of the range were rare and in particular, there were no forward skips with $\delta i>100$ in the performance data.
Skips with $-3\leq \delta i\leq 3$, which can be treated as insertion and deletion errors, are omitted from the figure.
Although the distribution clearly illustrates these features and indicates a somewhat continuous nature for $\delta i\lesssim -17$,
more extensive analysis with a larger data set is needed.
These features can be useful for further improving the following time and reducing the error rate of estimation, but since the features cannot be described by the outer-product HMM, we must return to the full HMM, which suffers from large computational complexity, to implement the features into the performance model.
An algorithmic device for solving this problem is also needed.

Although the proposed algorithms have a relatively low error rate in estimating score positions for practical performances with mistakes and skips as they stand, further studies are desired for reducing the error rate.
For this purpose, ornaments should be treated with care, since performance uncertainty is high in them \cite{Dannenberg1988,Tekin2006} (see also Sec.\ \ref{sec:PerformanceUncertainty}).
Also, the use of temporal and voice information proposed, for example, in Ref.~\cite{Gingras2011} may be important for improvements to both online and offline algorithms, especially when the left and right hands are weakly synchronised.
For this to be achieved for online algorithms, an extended model and/or an improved algorithm are necessary.
We are currently working on these matters.

\section{Summary}\label{sec:Summary}
In this paper, we discussed the score following of polyphonic MIDI performances with arbitrary repeats and skips and performance mistakes based on probabilistic models of musical performances.
In order to solve the problem with large computational complexity, we proposed a new type of HMM with the transition probability matrix composed of a band matrix and an outer product of two vectors, and derived efficient inference algorithms with reduced computational complexity for the HMM.
The HMM can describe performers' tendencies in distributions of the stop and resumption positions, and we discussed how much such information would improve the results for score following if the distributions are less widespread in score positions.

By analysing performances by three pianists during practice, we found that the distributions, especially those of the resumption positions, are indeed less widespread, and confirmed that the score-following results improves through evaluation.
The evaluation results show that the score-following algorithm can follow arbitrary repeats and skips within about two chords on average, and the error rate is under about 10\% for performances with high rates of performance mistakes,
proving that the algorithm is effective in practical situations.
We also illustrated that the offline version of the score-performance matching algorithms yield much better results than the online score-following algorithms, and that the offline algorithms could be effectively used for estimating the distributions of stop and resumption positions.

\section*{Acknowledgements}
This work is supported in part by Grant-in-Aid for Scientific Research from
Japan Society for the Promotion of Science, No.~23240021 (S.S. and N.O.).

\appendix

\section{Inference algorithms for general outer-product\\ HMM}\label{app:InferenceAlgorithmForGeneralCase}

In Appendix \ref{app:InferenceAlgorithmForGeneralCase}, we derive efficient inference algorithms for  general outer-product HMMs with output probability given in Eq.~(\ref{eq:OuterProductOutputMatrix}). We also relax the condition $\alpha_{j,i}\geq0$ for all $i$ and $j$, which was assumed in Sec.~\ref{sec:InferenceAlg}.

We first derive a fast Viterbi algorithm.
The Viterbi update in Eq.~(\ref{eq:ViterbiUpdate}) is now appropriately generalised as
\begin{equation}\label{eq:OuterProductViterbiUpdateGeneral}
\hat{p}_M(i)=\max_{j}\left[\hat{p}_{M-1}(j)a_{j,i}b_{j,i}(o_M)\right].
\end{equation}
Substituting Eqs.\ (\ref{eq:OuterProductTrMatrix}) and (\ref{eq:OuterProductOutputMatrix}), we have
\begin{equation}
\hat{p}_M(i)
=\max\left\{\max_{j \in {\rm nbh}(i)}[\hat{p}_{M-1}(j)a_{j,i}\beta_{j,i}(o_M)],
r_iu_i(o_M)\max_{j \notin {\rm nbh}(i)}[\hat{p}_{M-1}(j)S_jv_j(o_M)]\right\}.
\label{eq:FastViterbiOuterProductGeneral}
\end{equation}
The computational complexity of the second term of Eq.\ (\ref{eq:FastViterbiOuterProductGeneral}) is naively ${\cal O}(N(N-D))$ because $N-D$ elements over $N$ states are calculated.
We remind ourselves that once the $D+1$ largest values in $\{\hat{p}_{M-1}(j)S_jv_j(o_M)\}_{j=1}^N$ are found, one of them is always the solution to the maximum of the second term in Eq.\ (\ref{eq:FastViterbiOuterProductGeneral}) for each $i$.
Since finding the $D+1$ largest values and calculating the maximum in the second term of Eq.\ (\ref{eq:FastViterbiOuterProductGeneral}) for each $i$ have only ${\cal O}(N)$ complexity for the former and ${\cal O}(D+1)$ complexity for the latter, we can reduce the total computational complexity to ${\cal O}((2D+1)N)$.
Note that space complexity increases compared to the previous case since we now need to store the $D+1$ largest values of $\{\hat{p}_{M-1}(j)S_jv_j(o_M)\}_{j=1}^N$; however, this does not cause serious delays in processing times, at least if $D$ is small.

We can also derive efficient algorithms similarly for the forward and backward algorithm.
The forward algorithm can be refined as
\begin{align}
F_m(i)&=\sum_jF_{m-1}(j)a_{j,i}b_{j,i}(o_m)\\
&=\sum_{j \in {\rm nbh}(i)}F_{m-1}(j)a_{j,i}\beta_{j,i}(o_m)+\sum_{j \notin {\rm nbh}(i)}F_{m-1}(j)S_jr_iv_j(o_m)u_i(o_m)\\
&=\sum_{j \in {\rm nbh}(i)}F_{m-1}(j)\left(a_{j,i}\beta_{j,i}(o_m)-S_jr_iv_j(o_m)u_i(o_m)\right)
+r_iu_i(o_m)\sum_{j}F_{m-1}(j)S_jv_j(o_m).
\label{eq:FastForwardAlgGeneral}
\end{align}
Since the last summation on $j$ can be computed independently of $i$, the computational complexity is reduced to ${\cal O}(DN)$.
The backward algorithm can also be written as
\begin{align}
B_{m-1}(i)&=\sum_ja_{i,j}b_{i,j}(o_m)B_m(j)\\
&=\sum_{j \in {\rm nbh}(i)}\left(a_{i,j}\beta_{i,j}(o_m)-S_ir_jv_i(o_m)u_j(o_m)\right)B_m(j)
+S_iv_i(o_m)\sum_jr_ju_j(o_m)B_m(j).
\end{align}

\section{Evaluation of score-position estimating algorithms with varying $D$}\label{app:DetailedAnalysis}

\begin{table}[tb]
\caption{Error rate (\%) for the play-through performances for $D=4$, 10 and 20. Same abbreviations as in Table \ref{tab:AccRareSkip} are used for the algorithms.}
\label{tab:AccRareSkipDetail}
\begin{center}
\begin{tabular}{l|c|c|c}\hline\hline
Algorithm & Debussy & Mozart & Mussorgsky\\
\hline
O-on ($D{=}4$)&4.56&4.63&1.58\\
O-on ($D{=}10$)&3.24&5.03&1.87\\
O-on ($D{=}20$)&3.11&5.10&1.87\\
\hline
U-on ($D{=}4$)&3.53&5.53&1.68\\
U-on ($D{=}10$)&2.78&5.00&1.93\\
U-on ($D{=}20$)&2.81&5.10&1.93\\
\hline
N-on ($D{=}4$)&2.58&4.56&1.17\\
N-on ($D{=}10$)&2.05&5.00&1.71\\
N-on ($D{=}20$)&2.44&5.09&1.79\\
\hline\hline
O-off ($D{=}4$)&0.79&2.87&0.62\\
O-off ($D{=}10$)&0.53&3.04&0.62\\
O-off ($D{=}20$)&0.50&2.80&0.72\\
\hline
U-off ($D{=}4$)&0.69&2.90&0.76\\
U-off ($D{=}10$)&0.46&2.83&0.76\\
U-off ($D{=}20$)&0.50&2.80&0.78\\
\hline
N-off ($D{=}4$)&0.73&2.94&0.58\\
N-off ($D{=}10$)&0.46&3.04&0.60\\
N-off ($D{=}20$)&0.50&2.80&0.70\\
\hline\hline
\end{tabular}
\end{center}
\end{table}
In Appendix \ref{app:DetailedAnalysis}, we present the results for evaluation of score-position estimating algorithms with varying $D$.
In general, performance HMMs with larger $D$ have larger descriptive capabilities and the accuracy of score-position estimates must also be better if the parameters are appropriately set.
However, as we discussed in Sec. \ref{sec:Model} and \ref{sec:EvalCompComplex}, the computational complexity increases proportionally to $D$, and it is important to know how the results of estimation change with different $D$.
We use the same performance data and algorithms in Sec.\ \ref{sec:ScorePositionEstEvaluation}, and here we compare the results with three different values for $D$: $D=4$, 10 and 20.
See Table \ref{tab:ListD1D2} for the corresponding range of transitions in terms of $D_1$ and $D_2$.

The results for the play-through performances are listed in Table \ref{tab:AccRareSkipDetail}.
Although we see from both cases that the error rate increases or decreases with increasing $D$, overall, the effect from varying $D$ is rather small.
This is because skips are rare in the play-through performances and almost all the transitions are within the range of $D=4$.
Since the values of the transition probability are set by using data that include both the play-through performances and the performances during practice, the results can be worse for larger $D$.
If the values are set so that they are more suited to the play-through performances, the error rate should be smaller for larger $D$, but the difference is expected to be small as reasoned above.

\begin{table}[tb]
\caption{Error rate (\%) and its statistical error for the performances during practice for $D=4$, 10 and 20. Same abbreviations as in Table \ref{tab:AccRareSkip} are used for the algorithms.}
\label{tab:AccWithSkipDetail}
\begin{center}
\begin{tabular}{l|c|c|c}\hline\hline
Algorithm & Debussy & Mozart & Mussorgsky\\
\hline
O-on ($D{=}4$)&$13.0\pm3.1$&$10.7\pm2.1$&$9.5\pm1.6$\\
O-on ($D{=}10$)&$10.1\pm2.2$&$9.3\pm1.8$&$8.7\pm1.4$\\
O-on ($D{=}20$)&$10.4\pm2.1$&$9.6\pm1.9$&$9.1\pm1.2$\\
\hline
U-on ($D{=}4$)&$18.1\pm3.6$&$14.7\pm2.4$&$10.4\pm1.6$\\
U-on ($D{=}10$)&$13.9\pm2.3$&$12.1\pm1.7$&$8.3\pm1.2$\\
U-on ($D{=}20$)&$12.7\pm1.8$&$10.4\pm1.6$&$8.4\pm1.1$\\
\hline
N-on ($D{=}4$)&$39.0\pm2.8$&$40.3\pm4.3$&$18.3\pm2.1$\\
N-on ($D{=}10$)&$21.3\pm3.1$&$23.1\pm3.6$&$8.3\pm1.1$\\
N-on ($D{=}20$)&$15.6\pm2.1$&$16.2\pm1.8$&$8.3\pm1.0$\\
\hline\hline
O-off ($D{=}4$)&$4.8\pm1.6$&$5.4\pm1.2$&$3.0\pm0.7$\\
O-off ($D{=}10$)&$4.7\pm1.7$&$4.6\pm1.1$&$1.9\pm0.5$\\
O-off ($D{=}20$)&$4.1\pm1.4$&$4.7\pm1.0$&$2.0\pm0.5$\\
\hline
U-off ($D{=}4$)&$6.0\pm1.9$&$5.8\pm1.2$&$2.3\pm0.5$\\
U-off ($D{=}10$)&$5.2\pm1.5$&$4.9\pm1.1$&$2.1\pm0.5$\\
U-off ($D{=}20$)&$4.4\pm1.4$&$4.8\pm1.0$&$2.1\pm0.5$\\
\hline
N-off ($D{=}4$)&$27.0\pm3.3$&$23.3\pm4.2$&$12.5\pm1.5$\\
N-off ($D{=}10$)&$13.7\pm2.8$&$11.9\pm2.0$&$3.7\pm0.9$\\
N-off ($D{=}20$)&$9.3\pm2.2$&$9.1\pm1.6$&$2.3\pm0.5$\\
\hline\hline
\end{tabular}
\end{center}
\end{table}
\begin{table}[tbp]
\caption{Following rate (FR), averaged following time (FT), together with its statistical error, and its standard deviation (SD) of the online algorithms for the performances during practice. The algorithms and conditions are the same as in Table \ref{tab:AccWithSkip}.}
\label{tab:FTWithSkipDetail}
\begin{center}
\begin{tabular}{c|ccc}\hline\hline
\multicolumn{1}{c|}{Algorithm}&\multicolumn{3}{c}{Debussy}\\\cline{2-4}
 & FR (\%) & FT [chord] & SD [chord] \\
 \hline
O-on ($D{=}4$)&$93.3$&$2.31\pm0.18$&$2.90$\\
O-on ($D{=}10$)&$93.7$&$2.15\pm0.16$&$2.55$\\
O-on ($D{=}20$)&$94.8$&$2.22\pm0.15$&$2.40$\\
\hline
U-on ($D{=}4$)&$91.3$&$3.50\pm0.25$&$4.04$\\
U-on ($D{=}10$)&$93.7$&$3.25\pm0.24$&$3.79$\\
U-on ($D{=}20$)&$94.0$&$2.94\pm0.22$&$3.46$\\
\hline
N-on ($D{=}4$)&$(48.0)$&$\geq 11.25\pm0.53$&$(8.49)$\\
N-on ($D{=}10$)&$(73.8)$&$\geq 6.29\pm0.49$&$(7.70)$\\
N-on ($D{=}20$)&$(83.7)$&$\geq 4.55\pm0.38$&$(6.10)$\\
\hline\hline
\multicolumn{1}{c|}{Algorithm}&\multicolumn{3}{c}{Mozart}\\\cline{2-4}
 & FR (\%) & FT [chord] & SD [chord] \\
 \hline
O-on ($D{=}4$)&98.4&$2.03\pm0.13$&2.43\\
O-on ($D{=}10$)&98.7&$2.24\pm0.10$&1.99\\
O-on ($D{=}20$)&98.4&$2.36\pm0.11$&2.21\\
\hline
U-on ($D{=}4$)&97.3&$3.53\pm0.16$&3.05\\
U-on ($D{=}10$)&97.9&$3.30\pm0.14$&2.66\\
U-on ($D{=}20$)&96.8&$3.07\pm0.14$&2.74\\
\hline
N-on ($D{=}4$)&(64.6)&$\geq 12.31\pm0.71$&(13.66)\\
N-on ($D{=}10$)&(85.5)&$\geq 7.73\pm0.65$&(12.55)\\
N-on ($D{=}20$)&(89.0)&$\geq 5.64\pm0.44$&(8.46)\\
\hline\hline
\multicolumn{1}{c|}{Algorithm}&\multicolumn{3}{c}{Mussorgsky}\\\cline{2-4}
 & FR (\%) & FT [chord] & SD [chord] \\
 \hline
O-on ($D{=}4$)&97.6&$1.80\pm0.20$&1.79\\
O-on ($D{=}10$)&100&$1.80\pm0.16$&1.50\\
O-on ($D{=}20$)&98.8&$1.80\pm0.17$&1.51\\
\hline
U-on ($D{=}4$)&97.6&$2.11\pm0.21$&1.96\\
U-on ($D{=}10$)&96.4&$1.95\pm0.19$&1.74\\
U-on ($D{=}20$)&98.8&$1.90\pm0.19$&1.74\\
\hline
N-on ($D{=}4$)&(61.4)&$\geq 7.89\pm0.56$&(5.06)\\
N-on ($D{=}10$)&(69.9)&$\geq 4.98\pm0.56$&(5.06)\\
N-on ($D{=}20$)&(81.9)&$\geq 3.67\pm0.47$&(4.24)\\
\hline\hline
\end{tabular}
\end{center}
\end{table}
The results for the performances during practice are summarised in Tables \ref{tab:AccWithSkipDetail} and \ref{tab:FTWithSkipDetail}.
Since the tendencies of the results were similar for the algorithm O and O($\bm s,\bm r|_{\rm offline}$) and those with cross validation, only the results for the algorithm O are shown for simplicity\footnote{See Sec.\ \ref{sec:ScorePositionEstEvaluation} for the abbreviations of the algorithms.}.
The error rates for the algorithms O and U vary moderately with different $D$, and in most cases, those for $D=10$ and $D=20$ are within the $1\sigma$ errors.
The same is also true for the averaged following time.
In contrast, the results for the algorithm N vary significantly with different $D$, clearly illustrating the fact that the algorithm cannot handle skips out of the range defined by $D$.
For the Mussorgsky's piece, the error rate for the algorithm N-on is similar for $D=10$ and $D=20$ and slightly smaller than that for the algorithm O-on, which is explained by the fact that there are a few large skips in the performance samples.
The effectiveness of the algorithms O and U, or that of the algorithm N with larger $D$ for larger skips is manifested in the results for the averaged following time, in which the algorithms O and U outperform the algorithm N; the $D=20$ case outperforms the $D=10$ case for the algorithm N.

\end{document}